\documentclass{article}
\pdfoutput=1

\usepackage{bm}

\newcommand{\llb}[1]{{#1}}
\usepackage{graphicx}
\usepackage{multirow}
\usepackage{amsmath}
\usepackage{mathrsfs}

\usepackage{xspace}

\newcommand{\keep}[1]{}
\newcommand{\old}[1]{}

\DeclareMathOperator*{\softmax}{softmax}

\usepackage{times}
\usepackage{epsfig}
\usepackage{microtype}
\usepackage{graphicx}
\usepackage{subfigure}
\usepackage{booktabs} 

\usepackage{hyperref}

\usepackage[accepted]{icml2021}

\usepackage{cases, enumerate}

\begin{document}
\twocolumn[

\icmltitle{Unsupervised Co-part Segmentation through Assembly}



\icmlsetsymbol{equal}{*}

\begin{icmlauthorlist}
\icmlauthor{Qingzhe Gao}{sdu,bfa}
\icmlauthor{Bin Wang}{bfa}
\icmlauthor{Libin Liu}{pku}
\icmlauthor{Baoquan Chen}{pku}
\end{icmlauthorlist}

\icmlaffiliation{sdu}{Shandong University, Qingdao, Shandong, China}
\icmlaffiliation{bfa}{AICFVE, Beijing Film Academy, Beijing, China}
\icmlaffiliation{pku}{CFCS, Peking University, Beijing, China}

\icmlcorrespondingauthor{Baoquan Chen}{baoquan@pku.edu.cn}

\icmlkeywords{Machine Learning, ICML, Unsupervised Learning, Image Segmentation, Co-part Segmentation}

\vskip 0.3in
]



\printAffiliationsAndNotice{ }  

\begin{abstract}
Co-part segmentation is an important problem in computer vision for its rich applications. We propose an unsupervised learning approach for co-part segmentation from images. For the training stage, we leverage motion information embedded in videos and explicitly extract latent representations to segment meaningful object parts. More importantly, we introduce a dual procedure of part-assembly to form a closed loop with part-segmentation, enabling an effective self-supervision. We demonstrate the effectiveness of our approach with a host of extensive experiments, ranging from human bodies, hands, quadruped, and robot arms. We show that our approach can achieve meaningful and compact part segmentation, outperforming state-of-the-art approaches on diverse benchmarks.

\end{abstract}

\section{Introduction}\label{sec:intro}

Part-structure provides a compact and meaningful intermediate shape representation of articulated objects. Co-part segmentation, which aims to label semantic part belonging for each pixel of the objects in an image, is an important problem in computer vision. Such capability can directly serve various higher-level tasks such as marker-less motion tracking, action recognition and prediction, robot manipulation, and human-machine interaction.

 With the advent of deep learning, and the availability of large amount of annotated motion datasets, supervised learning-based approaches have led to superior performances over traditional part segmentation methods; the most success has been achieved for human pose estimation, e.g., ~\cite{guler2018densepose, kanazawa2018end}. However, this approach assumes significant domain knowledge, and highly depends on the specific dataset used for training, thus making it difficult to generalize to objects with different appearance, lighting or pose, not to mention unseen subjects. 

A video sequence is viewed as a spatio-temporal intensity volume that contains all structural and motion information of the action, including poses of the subject at any time as well as the dynamic transitions between the poses. Our goal in this research is to extract a general part-based representation from videos. Compared with single image-based segmentation, our work intends to aggregate shape correlation information from multiple images to improve the segmentation of individual images. The capability of consistently detecting object parts are important for motion tracking of creatures of various topology, and ultimately extracting their skeletal structures. 

A successful line of recent works in this direction formulates the task as an image generation problem, where segmented parts are globally warped to form the final image. There, part-segmentation becomes the essential intermediate step, because: the better you can segment (parts), the better you can generate (the image). In this paper, we follow the same image-generation concept, but introduce a dual procedure of part-assembly to form a closed loop with part-segmentation, which ensures more consistent, also more compact and meaningful part segmentation. Specifically, we generate the final image through blending each part's warped image, instead of a global image warping. In essence, our image based assembly operation effectively constrains the manifold of each individual part, resulting in improved results.  

We take an unsupervised learning approach. Like many recent works about unsupervised / self-supervised part segmentation, we believe shape correlation information between different frames can be leveraged for achieving semantic consistency. Our approach is similar to PSD~\cite{psd}, Motion Co-part~\cite{lathuiliere2020motion} and Flow Capsule~\cite{sabour2020unsupervised}, in the use of motion cues embedded in different frames for co-part segmentation. We go beyond the existing techniques in multiple ways:
\begin{enumerate}[(1)]
\setlength{\itemsep}{0pt}
	\item In our method, the supervision is attained by introducing a novel dual-procedure of part-assembly to form a close loop with part-segmentation.  
	\item The learned parts and their transformation have clear explainable physical meaning. 
	\item Our method doesn't require any field-based global warping operation, which enables handling dramatically dynamic motions. 
\end{enumerate}

We demonstrate the advantages of our method both visually and quantitatively. Extensive experiments have been conducted on datasets showcasing challenges due to the change in appearance, occlusions, scale and background. 
We also compared with recent works including NEM~\cite{NIPS2017_nem}, R-NEM~\cite{steenkiste2018relational_RNEM}, PSD~\cite{psd}, SCOPS~\cite{hung2019scops}, Motion Co-part~\cite{lathuiliere2020motion} and Flow Capsule~\cite{sabour2020unsupervised}. Our method outperforms state-of-the-art methods in quantitative evaluation.


\section{Related work}\label{sec:related_work}
\paragraph{Part-based Representation}
In analyzing images, describing object as a collection of parts, each with a range of possible states, is a classical framework for learning an object representation in computer vision ~\cite{ross2006learning, nguyen2013learning}. The states can be computed based on different evidences, such as visual and semantic features~\cite{wang2015semantic},  geometric shape and its behavior under viewpoint changes~\cite{eslami2012generative} and object articulation~\cite{sun2011articulated}, resulting in a large variation of part partition. Our work performs motion-based part segmentation, where each part is constituted with a group of pixels moving together.

\paragraph{Motion-based Co-part Segmentation}
Motion-based Co-part segmentation has been an important problem in understanding and reconstructing dynamic scenes.
Articulated object can be naturally segmented as a group of rigid parts, if prior knowledge on underlying kinematic structure is known. However, this assumption does not hold in our problem setting. 
Most of the traditional computer vision technology recovers rigid part and kinematic structure by exploiting motion information, in particular, RGB image sequences with feature points tracked over time. There have been three main approaches: (i) motion segmentation and factorization~\cite{Yan08}, (ii) probabilistic graphical model~\cite{Ross10, Sturm11}, and (iii) cost function based optimization methods~\cite{Ochs14, Keuper15}. The work from Chang and Demiris~\cite{Chang18} achieved state-of-the-art performance on the reconstruction of articulated structures from 2D image sequences. There, the segmentation was executed on tracked key-points, rather than all pixels like in our approach; the method, however, is prone to image noise, occlusions, deformations and cannot deal with articulated structures of high complexity.

\paragraph{Unsupervised Co-part Segmentation}
 With the popularity of deep neural networks, motion part segmentation has achieved superior performance in domains where labeled data are abundant, such as faces~\cite{khan2015multi} and human bodies~\cite{guler2018densepose, kanazawa2018end}. 
Parts segmentation can also be learned in an entirely unsupervised fashion. 
Nonnegative Matrix Factorization (NMF) ~\cite{lee1999learning} learned features that exhibit sparse part-based representation of data to disentangle the hidden structure of data. \cite{collins2018deep} further proposed deep feature factorization (DFF) to estimate the common part segments in images through NMF. 
Leveraging on semantic consistency in an image collection of single object category, Hung et. al. \cite{hung2019scops} proposed a self-supervised network SCOPS to predict part segmentation based on the pre-trained CNN features. 
\cite{psd} proposed a deep model to discover object parts and the associated hierarchical structure and dynamical model from unlabeled videos. However, they assume that pre-computed motion information is available. 
\cite{lathuiliere2020motion} proposed a model to leverage motion information with the purpose of obtaining segments that correspond to group of pixels associated to object parts moving together. But the transformation between parts in different frame is not explainable, and the motion merge to flow rather than on each part.
\cite{sabour2020unsupervised} proposed exploit motion as a powerful perceptual cue for part definition, using an expressive decoder for part generation and layered image formation with occlusion. But they still rely on flow to warp image, instead of considering that each part has independent motion.


\begin{figure*}[htp]
    \centering
    \includegraphics[width=0.9\textwidth]{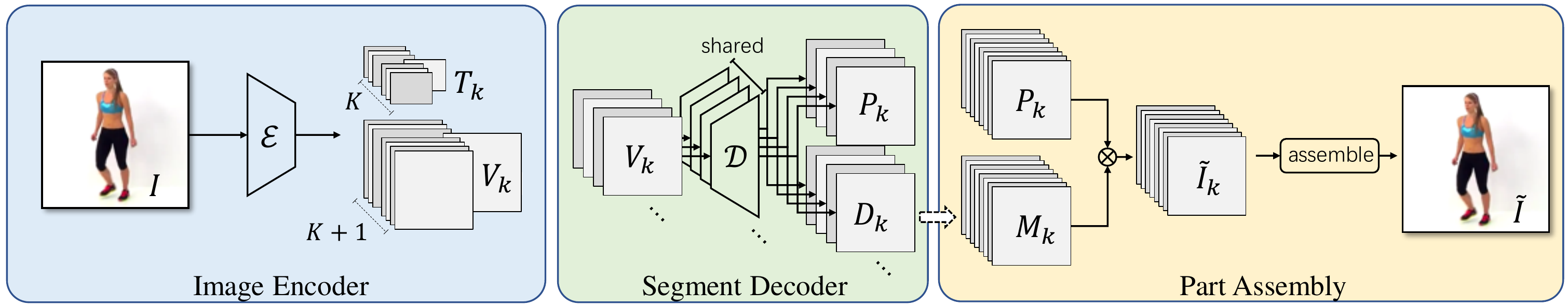}
    \caption{{
    \textbf{Architecture.} Our segmentation network consists of three major components. Left: The \emph{Image Encoder} takes an image $I$ as input and outputs latent feature maps $\{V_k\}$ and part transformations $\{T_k\}$ of every part. 
    Middle: The \emph{Segment Decoder} converts each feature map $V_k$ into a part image $P_k$ and a corresponding depth map $D_k$.
    Right: During the \emph{Part Assembly} procedure, the depth maps $\{D_k\}$ are converted into part masks $\{M_k\}$. The masked part images $\{\tilde{I}_k\}$ are then assembled to generate a reconstructed image $\tilde{I}$.
    }}
\label{fig:overview}
\vspace{-2mm}
\end{figure*}

\section{Method}\label{sec:method}

Our goal is to train a deep neural network to compute part segmentations and estimate part motions from a single input image.
We train our network using an image collection of the same object category, which can be extracted from videos of an animating object. 
Our unsupervised training process guides the network to identify object parts in the images by observing their motions.
To facilitate the training, we assemble the generated parts to recover the input images, which can be considered a dual-procedure of part segmentation. 
%
In what follows, we will first introduce our segmentation model in detail, then discuss the objectives and the training process.

\subsection{Model}\label{sec:method:model}

Our part segmentation network consists of an {\emph{image encoder}} network and a \emph{segment decoder} network. The image encoder encodes a given image into a set of latent feature maps, each corresponding to an object part. The segment decoder then decodes these part feature maps into part segments, which are later assembled to reconstruct the input image. Figure~\ref{fig:overview} provides an overview of the proposed network structure. 

\paragraph{Image Encoder}
Given an image $I\in\mathbf{R}^{H\times{}W\times{}C}$ as input, the image encoder, $\mathcal{E}$, computes latent representations of $K$ part segments, each represented by a feature map, $V_k\in{} \mathbf{R}^{H'\times{}W'\times{}C'}$, where $k\in\mathcal{K}$ and $\mathcal{K}=\{1,\dots,K\}$ is the set of part indices. 
%
%
These latent feature maps implicitly capture the shapes, appearances, locations, and poses of the corresponding part segments in the input image.
We treat the background region as a special part segment and represent the corresponding feature map as $V_0$.

The encoder also estimates a set of affine transformations, $\{T_k\}$, for each part $k\in\mathcal{K}$. We assume there exists a set of \emph{canonical} parts located at the center of the image, which are shared by all the images and can be transformed by $\{T_k\}$ to match the current parts in the input image. We use $V_k^*$ to represent the feature map of a canonical part.


In our implementation, a transformation, $T$, is given by a 6-tuple
\begin{equation}\label{eqn:transformation}
    T = (s_x, s_y, s_{\theta}, c_{\theta}, t_x, t_y)
\end{equation}
where $(s_x, s_y)$ and $(t_x, t_y)$ represent the scaling and the translation of the transformation. 
To avoid the continuity issue of angle representations \cite{zhou_continuity_2019}, we use two variables $(s_{\theta}, c_{\theta})$ to represent a rotation $\theta$, which correspond to the sine and cosine of $\theta$ respectively. The transformation matrix of $T$ is then given by
\begin{equation}
    T = \left[\begin{array}{cc}
         A & \bm{t}  \\
         \bm{0} & 1 
    \end{array} \right]
\end{equation}
where
\begin{equation*}
    A = 
    \left[\begin{array}{cc}
        \bar{c}_{\theta} & -\bar{s}_{\theta} \\
        \bar{s}_{\theta} & \bar{c}_{\theta} 
    \end{array}\right]
    \left[\begin{array}{cc}
        s_x & 0  \\
        0 & s_y 
    \end{array}\right],
    \bm{t} = \left[\begin{array}{c}
        t_x \\ t_y
    \end{array}\right]
\end{equation*}
and $(\bar{s}_{\theta},\bar{c}_{\theta}) = (s_{\theta}, c_{\theta})/\parallel(s_{\theta}, c_{\theta})\parallel_2$. 


\paragraph{Segment Decoder} 
The \emph{segment decoder} network, $\mathcal{D}$, is trained to convert a latent feature map ${V_k}$ into a part image, $P_k\in{}\mathbf{R}^{H\times{}W\times{}C}$, which recovers the appearance of the part in the original image.

We use the same decoder network $\mathcal{D}$ to convert feature maps of all the object parts and the background into part images. The decoder $\mathcal{D}$ also outputs a {depth map}, $D_k\in{}\mathbf{R}^{H\times{}W}$, for each part. With $\bm{u}$ representing the coordinates of a pixel, $D_k(\bm{u})$ is a scalar that specifies the relative inverse depth of the corresponding pixel located at $\bm{u}$ in part image $P_k$.
We assume that the object is composed of opaque parts, so that a part with smaller inverse depth (thus farther from the camera) will be partially occlude by the parts with larger inverse depth. 
%
%
The part mask $M_k\in\bm{R}^{H\times{}W}$ is thus a pixel-wise visibility mask indicating whether a pixel in the part image $P_k$ is visible in the original image, which can be computed as
\begin{equation}\label{eqn:part_mask}
    M_k(\bm{u})=\softmax\limits_{l\in{}\{0\} \cup{} \mathcal{K}} D_l(\bm{u})
\end{equation}

\paragraph{Part Assembly}
We train the segmentation network by assembling the part images together and reconstructing the input image. This is achieved by gathering the visible pixels from all the part images. Specifically, the reconstructed image $\tilde{I}$ is computed as
    $\tilde{I} = \sum_{k=0}^{K}\tilde{I}_k$
where $\tilde{I}_k$ is the visible part of part image $P_k$ as specified by the part mask $M_k$. We compute $\tilde{I}_k$ using
    $\tilde{I}_k = {M}_k \odot{} P_k$,
where $\odot$ is the Hadamard (pixel-wise) product between two arrays.

\begin{figure*}[htp]
    \centering
    \includegraphics[width=0.95\textwidth]{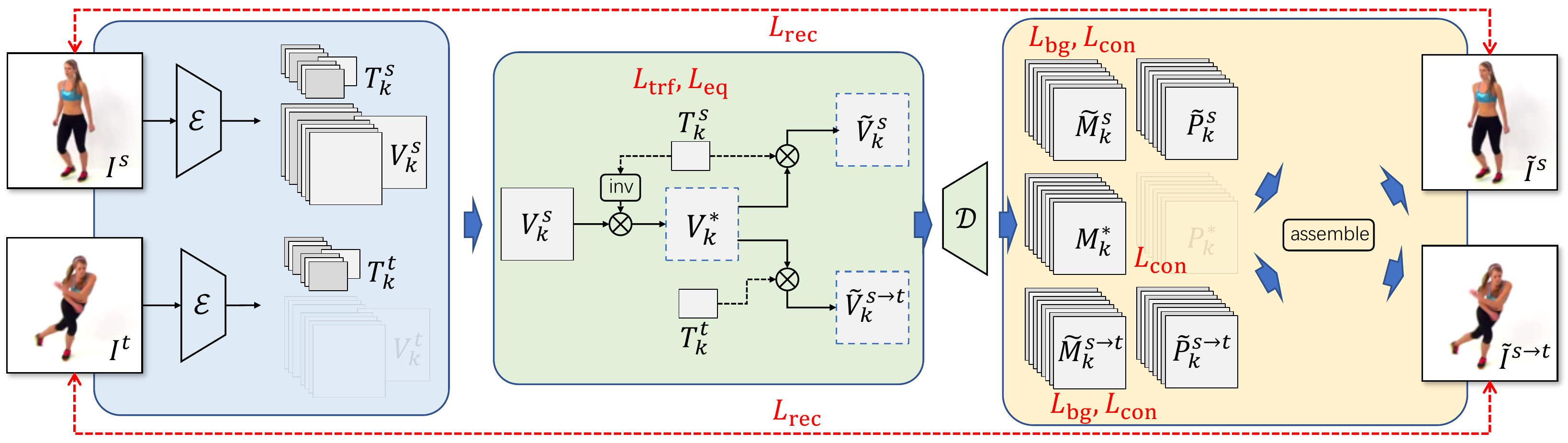}
    \caption{{
        \textbf{Training Process.} We train our segmentation network in an end-to-end fashion. 
        Left: The encoder $\mathcal{E}$ converts both the source image $I^s$ and the target image $I^t$ into latent feature maps and estimates part transformations.
        Middle: The source latent feature maps $V_k^s$ is inversely transformed into the canonical feature map $V_k^*$ using the source transformation $T_k^s$. $V_k^*$ is then transformed using $T_k^s$ and the target transformation $T_k^t$, producing a recovered feature map $\tilde{V}_k^s$ and a retargeted feature map $\tilde{V}_k^{s\rightarrow{}t}$ respectively.
        Right: The decoder $\mathcal{D}$ assembles the resulting $\{\tilde{V}_k^s\}$ and $\{\tilde{V}_k^{s\rightarrow{}t}\}$ and generates a reconstructed source image $\tilde{I}^s$ and a retargeted image $\tilde{I}^{s\rightarrow{}t}$ respectively. Those generated images are then compared with the input to compute losses.
    }}
\label{fig:training}
\vspace{-2mm}
\end{figure*}

\subsection{Training}\label{sec:training}
We train our co-part segmentation network using image pairs randomly selected from the input image collection. During training, we require the network to reconstruct one image of a pair (the \emph{source} image, $I^s$) as accurate as possible, while using the other image of the pair (the \emph{target} image, $I^t$) to cross-validate the latent representation of the parts and the segmentation results. 
This validation is performed by constructing the target image using the part segments extracted from the source image.
{Unlike the existing works that warp the source image using optical flow \cite{siarohin2020first,siarohin2020motion,sabour2020unsupervised}, we transform the parts in latent space directly and decode the transformed latent features of the parts to generate the target image.} 

Figure~\ref{fig:training} provides an overview of this training process. In more details, the encoder network $\mathcal{E}$ takes the two images $I^s$ and $I^t$ as input and computes latent feature maps and part transformations for both of them. The results are denoted as $\{V^s_0, (V_k^s, T_k^s)\}$ and $\{V^t_0, (V_k^t, T_k^t)\}$ respectively. 
We transform each source latent feature map $V_k^s$ using the corresponding transformations $T_k^s$ and $T_k^t$, and assemble the resulting latent maps $\{\tilde{V}^{s\rightarrow{}t}_0, \tilde{V}^{s\rightarrow{}t}_k\}$ using the segment decoder $\mathcal{D}$ to produce the retargeted image $\tilde{I}^{s\rightarrow{}t}$. Note that we assume the background is static and use $\tilde{V}^{s\rightarrow{}t}_0 = V^s_0$ in this transformation.

This retargeting is performed in two steps. First, we inversely transform $V_k^s$ using $T_k^s$ to compute the canonical latent feature map $V_k^*=({T_k^s})^{-1}\circ{}V_k^s$, which is assumed to be shared by both the source and the target. 
The transformation operation $\circ{}$ is defined as 
\begin{equation}
(T\circ{}V)(\bm{u}) = V(T^{-1}\bm{u})
\end{equation}
where $\bm{u}$ and its transformed counterpart $T^{-1}\bm{u}$ are both coordinates of pixels.
Then the target transformation $T_k^t$ is applied to $V_k^*$ to compute the retargeted feature map $\tilde{V}^{s\rightarrow{}t}_k={T_k^t}\circ{}{V_k^*}=[{T_k^t}(T_k^s)^{-1}]\circ{}{V_k^*}$. 
We also recover the source feature map from $V_k^*$ as $\tilde{V}^{s}_k={T_k^s}\circ{}{V_k^*}$. We find this additional procedure helpful in facilitating the training at the early stage of the process.

The resulting feature maps $\{\tilde{V}^{s}_0, \tilde{V}^{s}_k\}$ and $\{\tilde{V}^{s\rightarrow{}t}_0, \tilde{V}^{s\rightarrow{}t}_k\}$ are then input to the decoder $\mathcal{D}$ to assemble the reconstructed image $\tilde{I}^s$ and the retargeted images $\tilde{I}^{s\rightarrow{}t}$ respectively. 
In the meanwhile, the corresponding part masks $\{\tilde{M}^s_k\}$, $\{\tilde{M}^{s\rightarrow{}t}_k\}$, and $\{M^*_k\}$, computed using $\mathcal{D}$ and Equation~\eqref{eqn:part_mask}, are recorded as well, which are used as a part of the training objective as described below.
We train our segmentation network in an end-to-end fashion with an objective formulated as a weighted sum of several losses. 

\paragraph{Image Reconstruction Loss}
The main driving loss of our training is the image reconstruction loss, which penalizes the difference between the generated images and the corresponding inputs.
The difference between images is measured based on the perceptual loss of~\cite{johnson2016vggloss}, where a pretrained VGG-19 network~\cite{simonyan2015vgg} is used to extract features from the images for comparison. The difference between two images $I$ and $\tilde{I}$ is then computed as
\begin{equation}\label{eqn:rec_loss}
\begin{aligned}
    \mathcal{L}(I, \tilde{I}) &= {\lambda_1}\parallel{}I - \tilde{I}\parallel_1
     + {\lambda_2}\parallel \nabla{}{I} - \nabla{}{\tilde{I}} \parallel_1 \\
    &+{\lambda_3}\parallel \phi_{\text{vgg}}({I}) - \phi_{\text{vgg}}({\tilde{I}})\parallel_1
\end{aligned}
\end{equation}
where $\nabla\cdot$ computes image gradient {as suggested by~\cite{eigen_predicting_2015}}, and $\phi_{\text{vgg}}(\cdot)$ extracts VGG-19 features from the image.
The reconstruction loss is applied to both the reconstructed source image $\tilde{{I}^s}$ and the regargeted image $\tilde{I}^{s\rightarrow{}t}$. The total loss is thus
\begin{equation}
    \mathcal{L}_\text{rec} = \lambda_{s}\mathcal{L}(I^s, \tilde{I}^s) + \lambda_{t}\mathcal{L}(I^t, \tilde{I}^{s\rightarrow{}t})
\end{equation}


\begin{figure*}[t]
    \centering
   \newcommand{\formattedgraphics}[1]{\includegraphics[width=0.16\textwidth]{#1}}
   
    \formattedgraphics{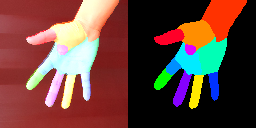}
    \formattedgraphics{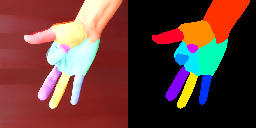}
    \formattedgraphics{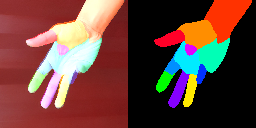}
    \formattedgraphics{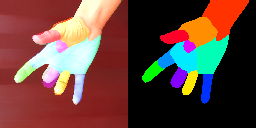}
    \formattedgraphics{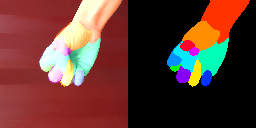}\\
    \formattedgraphics{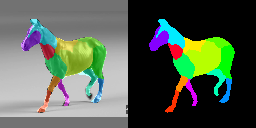}
    \formattedgraphics{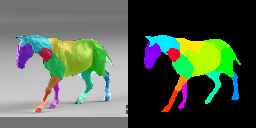}
    \formattedgraphics{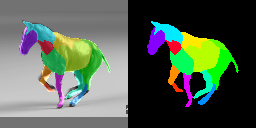}
    \formattedgraphics{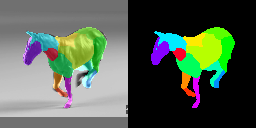}
    \formattedgraphics{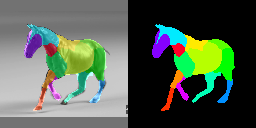}\\
    \formattedgraphics{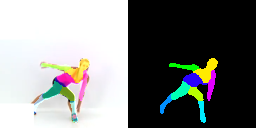}
    \formattedgraphics{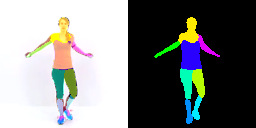}
    \formattedgraphics{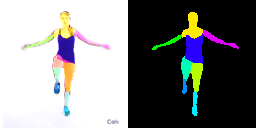}
    \formattedgraphics{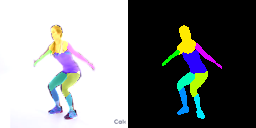}
    \formattedgraphics{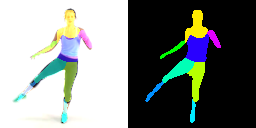}\\
    \formattedgraphics{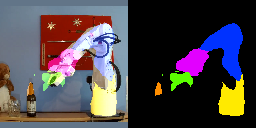}
    \formattedgraphics{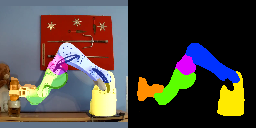}
    \formattedgraphics{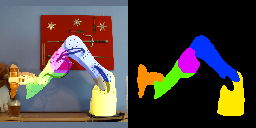}
    \formattedgraphics{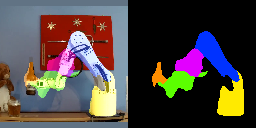}
    \formattedgraphics{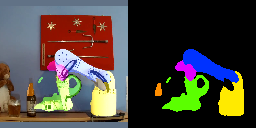}\\
    \vspace{-1mm}
    \caption{Visual results of our method tested on different scenarios, including human, hand, quadruped and robot arm. }
    \label{fig:other}
\vskip -0.2in
\end{figure*}

\paragraph{Background Loss}
As described in the last section, we treat the background as a special part segment and compute its features map and part mask. In practice, however, we observe that some background pixels can appear in the other part masks, causing noisy part segmentation.
To address this problem, we include a novel background loss in the training to encourage clear partition between the object parts and the background.
%
%
This is achieved by encouraging the background part to occupy as much image as possible, forcing the object parts to shrink into the most relevant region in the image.
The background loss is thus defined as
\begin{equation}
\mathcal{L}_{\text{bg}} = \lambda_{\text{bg}}
    \left(\parallel M_0 - 1 \parallel_1 + \sum_{k=1}^{K} \parallel {M_k}\parallel_1 \right)
\end{equation}
which drives the values of the background mask close to one and the values of the other part masks close to zero.
We apply this background loss to both the source parts $M^s_k$ and the retargeted parts $M^{s\rightarrow{}t}_k$. We find this loss essential for precise part segmentation with tight boundaries.

\paragraph{Transformation Loss}
%
In our system, we expect each part transformation $T_k$ to be strongly correlated with the absolute pose of the part $k$ in the input image and thus has a clearly explainable physical meaning. 
More specifically, each part transformation $T_k= \left[A_k | \bm{t}_k \right]$ defines a coordinate system with the origin at $\bm{t}_k$ and the axes defined by $A_k$. We assume $\bm{t}_k$ to be located at the center of the part $k$ and the axis $A_k$ align with the longest and the shortest dimensions of the object part. We enforce such property in the training using a novel transformation loss defined as $\mathcal{L}_\text{trf}=\mathcal{L}_\text{tran}+\mathcal{L}_\text{rots}$, where
\begin{align}
    \mathcal{L}_\text{tran}&=
    \lambda_\text{tran}\sum_{k=1}^K\parallel \bm{t}_k - \bm{\hat{u}}_k \parallel_1 \\
    \mathcal{L}_\text{rots}&=
    \lambda_\text{rots}\sum_{k=1}^K\parallel A_k A_k^T - \Sigma_k \parallel_F
\end{align}
%
%
%
%
Since we do not have the ground-truth part poses, we estimate the reference transformation using the mean and the covariance of the part mask $M_k$ as 
\begin{align}
    \bm{\hat{u}}_k&=\frac{1}{z_k}{\sum_{\bm{u}\in{}\mathcal{U}}\bm{u}{M_k}(\bm{u})} \label{eq:COM}\\
    \Sigma_k&=\frac{1}{z_k} {\sum_{\bm{u}\in{}\mathcal{U}}
        (\bm{u}-\bm{\hat{u}}_k)(\bm{u}-\bm{\hat{u}}_k)^T{M_k}(\bm{u})}
    \label{eq:part_cov}
\end{align}
where $M_k(\bm{u})$ represents the mask value of the pixel located at $\bm{u}\in\mathcal{U}$ and $z_k=\sum_{\bm{u}\in{}\mathcal{U}}{M_k}(\bm{u})$ is a normalization constant. In this estimation, we only consider the pixels that have been clearly identified as part $k$ with a threshold $\zeta$, so that $\mathcal{U}=\{\bm{u}|M_k(\bm{u})>\zeta\}$. {We choose $\zeta=0.02$ empirically in our implementation.}
%
%

\begin{figure*}[t]
    \centering
   \newcommand{\formattedgraphics}[1]{\includegraphics[width=0.9\textwidth]{#1}}
    \formattedgraphics{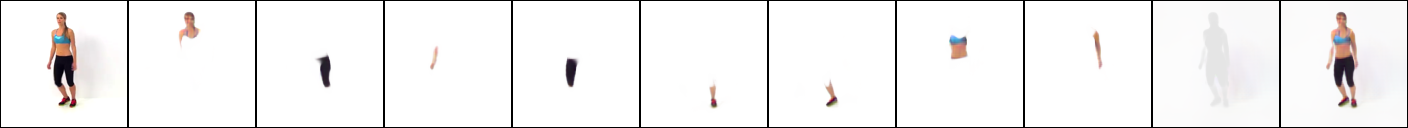}\\
    \formattedgraphics{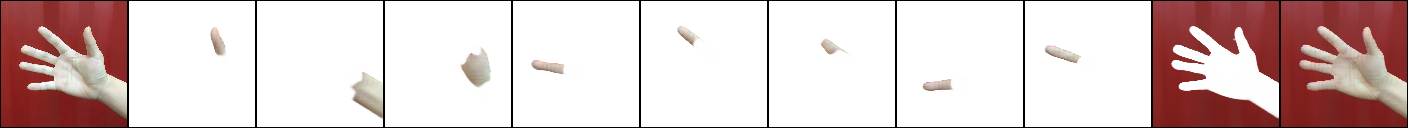}\\
    \formattedgraphics{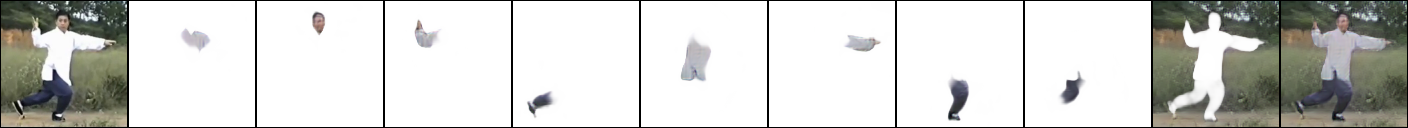}\\
    \caption{Visualization of the segmentation results. Left: The input image. Middle: The part images computed by our segment decoder. Right: The reconstructed image computed by part assembly.}
    \label{fig:vis_part_image}
\vskip -0.2in
\end{figure*}

\paragraph{Equivariance Loss}
The estimation of the part transformations should be consistent across images and show equivariance to image transformations. Following the common practice of unsupervised landmark detection~\cite{jakab_unsupervised_2018,zhang_unsupervised_2018,siarohin2020first}, we employ an equivariance loss in our training. 
Specifically, we transform the input image $I$ using a random transformation $T^w$. The encoder $\mathcal{E}$ then estimates part transformations for both the original image $I$ and the transformed image $I^w$. 
The equivariance loss is then defined as
\begin{equation}\label{eqn:eq_loss}
    \mathcal{L}_{\text{eq}} = \lambda_{\text{eq}} \sum_{k=1}^K \parallel T^wT_k - T^w_k \parallel_1
\end{equation}
where $T_k$ and $T^w_k$ are the part transformations estimated from $I$ and $I^w$ respectively, and the 1-norm is computed with the transformation matrix treated as a vector.


\paragraph{Concentration Loss}
To encourage the pixels belonging to the same object part to form a connected and concentrated component,
we employ the geometry concentration loss suggested in SCOPS~\cite{hung2019scops} to regularize the shape of part mask $M_k$. Specially, this concentration loss is computed as
\begin{equation}\label{eqn:con_loss}
    \mathcal{L}_{\text{con}} =\lambda_{\text{con}} \sum_{k=1}^{K} \sum_{\bm{u}} \parallel \bm{u}  - \bm{\bar{u}}_k \parallel^2_2 \cdot {M_k}(\bm{u})/{z_k}
\end{equation}
where 
\begin{equation}\label{eqn:com_all}
    \bm{\bar{u}}_k=\frac{1}{z_k}\sum_{\bm{u}}\bm{u}{M_k}(\bm{u})
\end{equation}
is the center of gravity of the part mask $M_k$, and $z_k=\sum_{\bm{u}}{M_k}(\bm{u})$ is a normalization constant. Unlike the transformation loss, we consider all the pixels in $M_k$ in this concentration loss. 
Note that the summation of Equation~\eqref{eqn:con_loss} excludes the part mask corresponding to the background.
As shown in Fig.~\ref{fig:training}, we apply concentration loss to the part masks of the source parts $M^s_k$, the retargeted parts $M^{s\rightarrow{}t}_k$, and the {canonical} parts $M^*_k$.

\section{Experiments}\label{sec:experiments}

Our model is implemented using the the standard U-Net architecture. We include the details about the network structure and the training settings in the supplementary materials.
We visually demonstrate the effectiveness of our co-part segmentation method on several test cases with large variation, including human, hand, quadruped, and robot arms in Figure~\ref{fig:other}, where the resulting part segments are rendered with different colors
\llb{with the corresponding masks computed using \emph{hard}max $M_k(\bm{u})=\max_{l\in{0}\cup{} \mathcal{K}}D_l(\bm{u})$ instead of Equation~\eqref{eqn:part_mask}}.
Additionally, Figure~\ref{fig:vis_part_image} illustrates the individual segment images computed by our network.  
%

\llb{
Our method is designed to extract parts that exhibit different affine transformations in the training image pairs, which is consistent with the behavior of semantically meaningful segmentation of a subject, such as a human body.
%
The order of these parts is not determined in the unsupervised learning process. As in previous works, we manually label those parts after the model training. Notably, we only need to label the parts once, and the labels are consistent over all test images.
}

In the rest part of this section, we will introduce the ablation studies we performed to analyze the effectiveness of each loss component in our framework, and also the comparison with state-of-the-art co-part segmentation techniques. 

\begin{figure*}[t]
   \vspace{-2mm}
    \centering
   \newcommand{\formattedgraphics}[1]{\includegraphics[width=0.95\textwidth]{#1}}
   
    \formattedgraphics{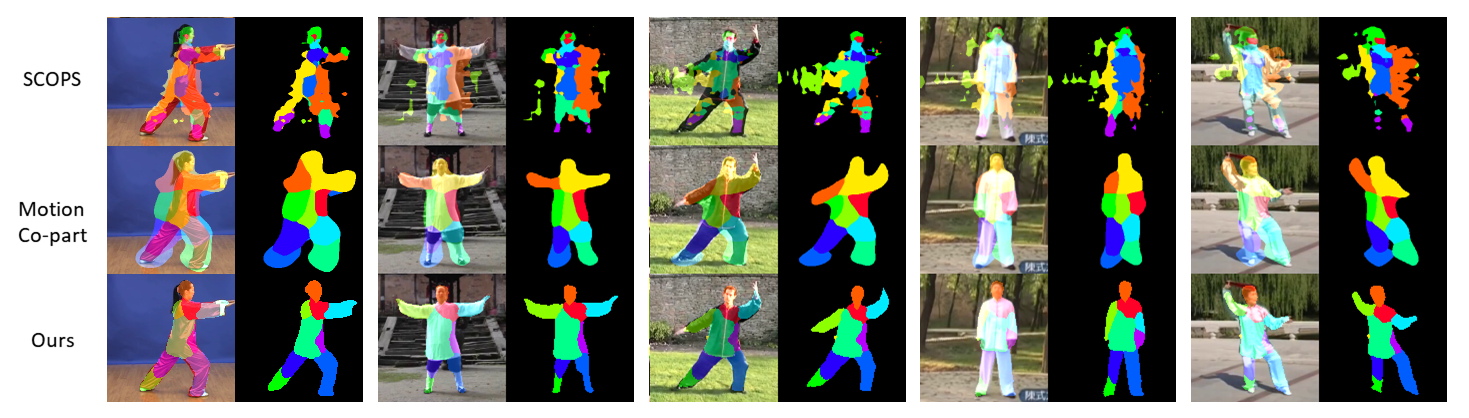}
    \vspace{-2mm}
    \caption{Visual result in Tai-Chi-HD. Motion Co-part and our method both produce relatively better and consistent part segmentation than SCOPS. Our segmentation are more compact and tightly aligned with image silhouette.}
    \label{fig:taichi}
    \vspace{-2mm}
\end{figure*}

\begin{figure*}[t]
    \centering
   \newcommand{\formattedgraphics}[1]{\includegraphics[width=0.95\textwidth]{#1}}
   
    \formattedgraphics{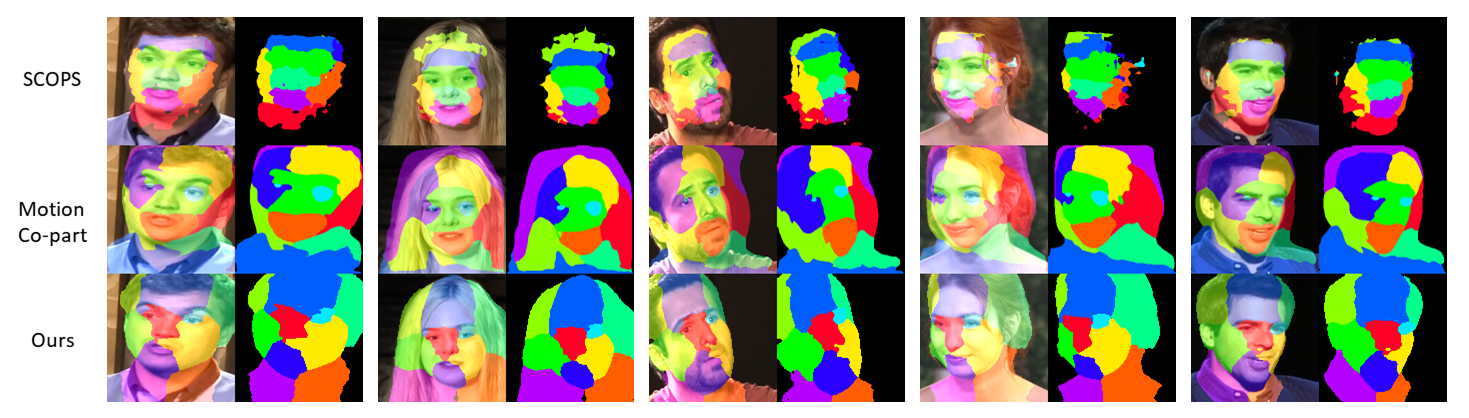}
    \vspace{-2mm}
    \caption{Visual result on VoxCeleb. All of the three methods produce consistent part segments, but our segmentation are more compact and tightly aligned with the image silhouette.}
    \label{fig:VoxCeleb}
    \vspace{-4mm}
\end{figure*}


\begin{table}[t]
\caption{The quantitative evaluation on the validation set of Tai-Chi-HD and VoxCeleb. The evaluation
metrics are the foreground IOU and landmark regression MAE. 
}
\label{taichi}
\vskip 0.15in
\begin{center}
\begin{small}
\begin{tabular}{ccccc}
\toprule
Dataset & Metric & SCOPS & Motion & Ours  \\
  &  &  & Co-part  &   \\
\midrule
\multirow{2}{*}{Tai-Chi-HD} & Landmark & 411.38 & 389.78 & \bf{326.82}\\
 & IoU         & 0.5485 & 0.7686 & \bf{0.8724}\\
 \midrule
\multirow{2}{*}{VoxCeleb} & Landmark   & 663.04 & 424.96 & \bf{338.98}\\
& IoU  & 0.5045 & 0.9135 & \bf{0.9270}\\
\bottomrule
\end{tabular}
\end{small}
\end{center}
\vskip -0.1in
\vspace{-2mm}
\end{table}

\subsection{Datasets}
\paragraph{Tai-Chi-HD.} Tai-Chi-HD dataset~\cite{siarohin2020first} is a collection of short videos with full-body Tai-Chi movements. $2981$ Tai-Chi videos were downloaded from YouTube. The videos were cropped and resized to a fixed resolution of $128 \times 128$, while preserving the aspect ratio. There are $2746$ training videos and $235$ test videos. This dataset contains $5300$ images with ground truth landmarks ($18$ joints) generated using the method from Cao et al.\cite{cao2019openpose}. Only $300$ images with ground truth foreground segmentation mask are available.

\begin{table}[t]
\caption{The quantitative evaluation on the validation set of Exercise. The evaluation metric is the part segment IoU.}
\label{table:yoga}

\vskip 0.15in
\begin{small}
\begin{tabular}{cccccc}
\toprule
Part&REM&N-REM&PSD &Flow &Ours\\
& & &  & Capsule&\\
\midrule
Full   & 0.298& 0.321  &0.697  & -  &\bf{0.793}   \\
Upper  & 0.347& 0.319  &0.574  & 0.690  & \bf{0.759} \\  
Arm     & 0.125& 0.220  &0.391  & -     & \bf{0.465}  \\
Leg(L)  & 0.264& 0.294  &0.374  & 0.590 & \bf{0.726}   \\ 
Leg(R)  & 0.222& 0.228  &0.336  & 0.540 &  \bf{0.642}   \\    
Average   & 0.251& 0.276  &0.474  & - &   \bf{0.677} \\        
\bottomrule
\end{tabular}
\end{small}
\vskip -0.1in
\end{table}

\paragraph{VoxCeleb.} The VoxCeleb dataset~\cite{nagrani2017voxceleb} is a large scale face dataset, which consists of $22496$ videos, extracted from YouTube. We follow the preprocessing described in ~\cite{siarohin2020first} to crop original video into several short sequences to guarantee that face can move freely in the image space with reasonable scale. All the cropped videos are then resized to $128 \times 128$, again, preserving the aspect ratio. After the preprocessing, our dataset contains $15103$ training videos and $443$ test videos. The length of each video varies from $64$ to $1024$ frames. 
This dataset contains $5300$ images with ground truth landmarks ($68$ keypoints) generated using the method from Bulat et al.\cite{bulat2017far}. Similarly, only $300$ images with ground truth foreground segmentation mask are available.

\paragraph{Exercise.} Exercise dataset is a collection of paired images from two consecutive frames for full human body performing Yoga exercises. This dataset is originally collected by \cite{xue2016visual} from YouTube, and preprocessed with motion stabilization. We use the reorganized version of this dataset provided by \cite{psd}, which contains $49356$ pairs of images for training. For this dataset, only $30$ images with ground truth part segmentation masks are available.


\subsection{Metrics}
\paragraph{Intersection Over Union}
We use the commonly adopted mean intersection over union (IoU) metric to evaluate how similar our predicted segmentation is to the ground truth. The average IoU across all frames of the dataset is practically used. 
We use foreground IoU for the test on VoxCeleb and Tai-Chi-HD datasets due to the shortage of ground-truth part segmentation mask; while part IoU is used for the test on Exercise dataset.
\paragraph{Landmark Regression MAE}
We use the landmark regression MAE metric to evaluate whether our method can generate consistent semantic part segmentation on different images. Following 
~\cite{hung2019scops}, We first fit a linear regression model from the parts' center of mass $\bm{\bar{u}}_k$ to ground truth landmarks using $5000$ annotated images, where $\bm{\bar{u}}_k$ is calculated using Equation~\ref{eqn:com_all}, and then on the other $300$ images, we compute the mean average error (MAE) between regressed and ground truth landmark positions as the evaluation metric.

\vspace{-0.25cm}
\begin{figure*}[h]
\centering
   \newcommand{\formattedgraphics}[1]{\includegraphics[width=0.14\textwidth]{#1}}
    \formattedgraphics{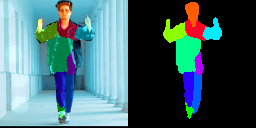}
    \formattedgraphics{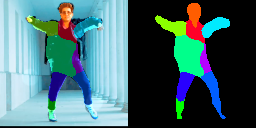}
    \formattedgraphics{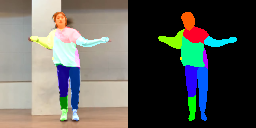}
    \formattedgraphics{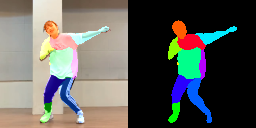}
    \formattedgraphics{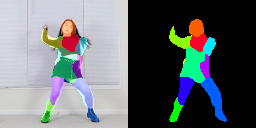}
    \formattedgraphics{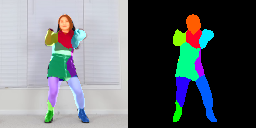}\\
    \formattedgraphics{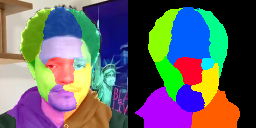}
    \formattedgraphics{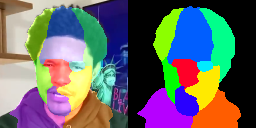}
    \formattedgraphics{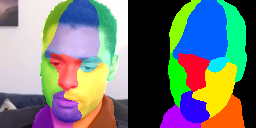}
    \formattedgraphics{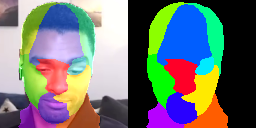}
    \formattedgraphics{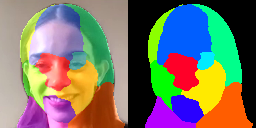}
    \formattedgraphics{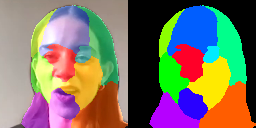}\\
    \formattedgraphics{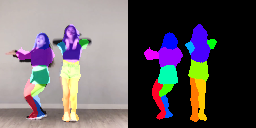}
    \formattedgraphics{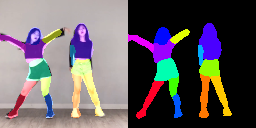}
    \formattedgraphics{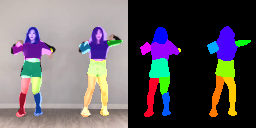}
    \formattedgraphics{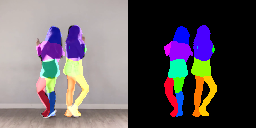}
    \formattedgraphics{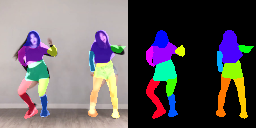}
    \formattedgraphics{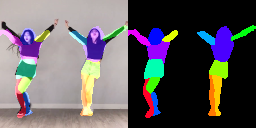}\\
    \vspace{-2mm}
    \caption{\textbf{Top, Middle}: Visual results of applying models trained on Tai-Chi-HD (top) and VoxCeleb (middle) to videos from YouTube.
    \textbf{Bottom}: Visual results of the model trained on a video with multiple characters.} 
    \label{fig:generalization}
\end{figure*}

\subsection{Quantitative Comparison}
We quantitatively compare our result with that of the state of the art methods for co-part segmentation, SCOPS~\cite{hung2019scops}, Motion Co-part~\cite{siarohin2020motion}, PSD~\cite{psd}, NEM~\cite{NIPS2017_nem}, N-REM~\cite{steenkiste2018relational_RNEM} and Flow Capsules~\cite{sabour2020unsupervised}, using both IoU and Landmark metrics. 


We first compared our method with SCOPS and Motion Co-part using landmark regression and foreground IoU metrics on VoxCeleb and Tai-Chi-HD dataset, which have rich variation on background texture, actor's appearance and body proportions, etc. To make a fair comparison, we train our model with $K=10$ in this test, which is consistent with the settings of the other methods. 
The results in Table~\ref{taichi} indicate that our method significantly improves the accuracy of foreground segmentation and can achieve more consistent and precise segmentation. As illustrated in Figures~\ref{fig:taichi} and \ref{fig:VoxCeleb}, our approach achieves more consistent part segmentation than SCOPS, where the main objects are clearly separated from the background. In the meanwhile, our results are aligned with image silhouette more tightly than Motion~Co-part.

We further compare our method with PSD, NEM, R-NEM and Flow Capsule on the accuracy of co-part segmentation using the IoU metric on Exercise dataset. We train our model with $K=15$ segments in this comparison. 
As reported in Table~\ref{table:yoga}, 
our model achieves a consistently better 
performance than the baselines.



\subsection{Generalization}

\llb{Our model can be trained with both single-video settings (hand, quadruped, and robot arm) and multiple-video settings (Tai-Chi-HD, VoxCeleb, and Exercise datasets), where in the latter case, each pair of images are extracted from the same random video during training. 
Models trained with multiple videos can be generalized to images of the same category but with difference appearance. For example, without further training, the models trained on Tai-Chi-HD and VoxCeleb datasets can be applied directly to videos downloaded from YouTube, as shown in the top two rows in Figure~\ref{fig:generalization}.
}

\llb{
Moreover, it is rather straightforward to extend our method to support multiple subjects. The bottom row of Figure~\ref{fig:generalization} demonstrates an example of such ability, where a model is trained on a video with two persons, and $K = 17$ is used to accommodate additional potential part segments.
}

\begin{table}[t]
\vspace{-3mm}
\caption{Ablation study for different loss on Tai-Chi-HD.}
\label{ablation_study}
\vskip 0.15in
\begin{small}
\resizebox{0.98\columnwidth}{!}{
\begin{tabular}{ccccccc}
\toprule

Measures&w/o&w/o&w/o& w/o&w/o&Full \\
&$\mathcal{L}_\text{vgg}$ & $\mathcal{L}_\text{bg}$  & $\mathcal{L}_\text{rots}$& $\mathcal{L}_\text{tran}$&$\mathcal{L}_\text{con}$&\\
\midrule
Landmark & 386.1 & 350.6 & 335.8 & 334.5 & 366.6 & 326.8 \\
 IoU     & 0.784 & 0.828 & 0.856 & 0.861 & 0.861 & 0.872\\
\bottomrule
\end{tabular}
}
\end{small}
\vskip -0.1in
\vspace{-2mm}
\end{table}

\subsection{Ablation Study}
\begin{figure}[t]
    \centering

    \includegraphics[width=0.9\columnwidth]{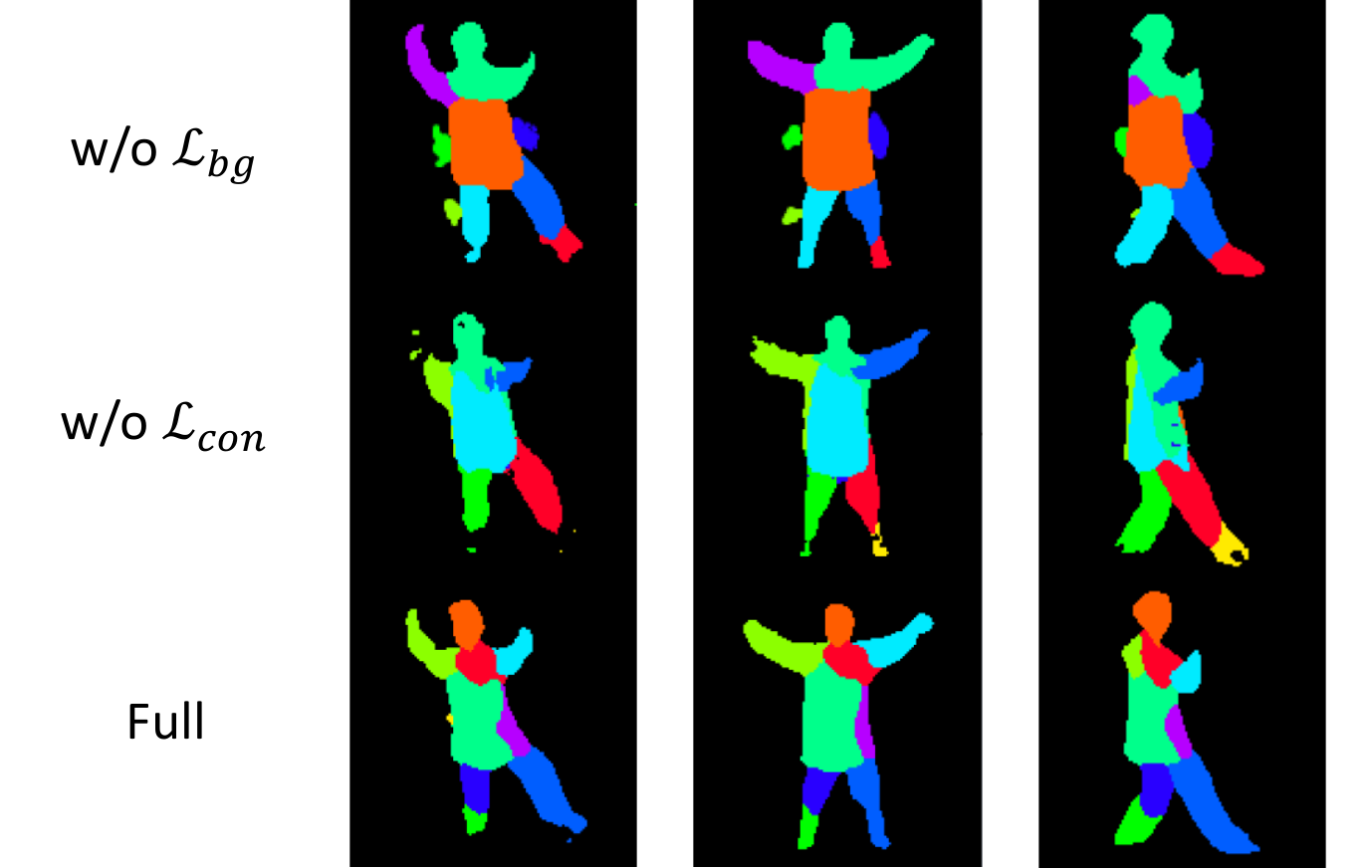}
    \caption{Visual comparison of segmentation results by using full loss, disable ${\cal L}_{con}$ and ${\cal L}_{bg}$ individually. Without ${\cal L}_{bg}$, background information distributed into many channels; while without ${\cal L}_{bg}$, the foreground segmentation have severe noisy. }
    \label{fig:difloss}
    \vspace{-2mm}
\end{figure}

We perform ablation studies to
validate the contribution of each loss employed in our training. The comparison is conducted between the full training objective proposed in Sec.\ref{sec:training} and its variants, each with one of the losses ($\mathcal{L}_\text{vgg}$, $\mathcal{L}_\text{bg}$,  $\mathcal{L}_\text{rots}$, $\mathcal{L}_\text{tran}$, $\mathcal{L}_\text{con}$) disabled.
\llb{We use Tai-Chi-HD dataset in these experiments, and the results are reported in terms of foreground IoU and landmark regression accuracy.}
Note that in the ablation study with $\mathcal{L}_\text{vgg}$, we only remove the VGG features from $\mathcal{L}_\text{rec}$ and keep the rest terms of Equation~\eqref{eqn:rec_loss} unchanged.
The results summarized in Table~\ref{ablation_study} reveal that all the losses are beneficial towards effective learning. 

$\mathcal{L}_\text{vgg}$ is the most significant one among them, which influences both foreground extraction and part segmentation. 
Unlike the methods that generate images using field-based global warping operation~\cite{siarohin2020first,sabour2020unsupervised}, our model cannot utilize the pixels of the input image directly to generate the target image, where the VGG features significantly facilitate the training and help achieve a good performance.
%
%

$\mathcal{L}_\text{bg}$ is another critical term of our objective design, which enforces all the background information to be embedded in the background channel, thus consequently ensuring the segmentation tightly aligned with the foreground silhouette. As shown in Figure~\ref{fig:difloss}, the segmentation trained without this loss can be noisy with background pixels mislabeled as a part of object segments.

Similar to the results reported in~\cite{hung2019scops}, we find $\mathcal{L}_\text{con}$ guarantees the semantic correctness of the segmented parts through penalizing vague and scattered partition in each channel. This conforms to our ablation study that $\mathcal{L}_\text{con}$ has obvious effect on Landmark regression accuracy. More visual comparison result can be found in Figure~\ref{fig:difloss}.  

\begin{figure}[t]
    \centering
   \includegraphics[width=0.9\columnwidth]{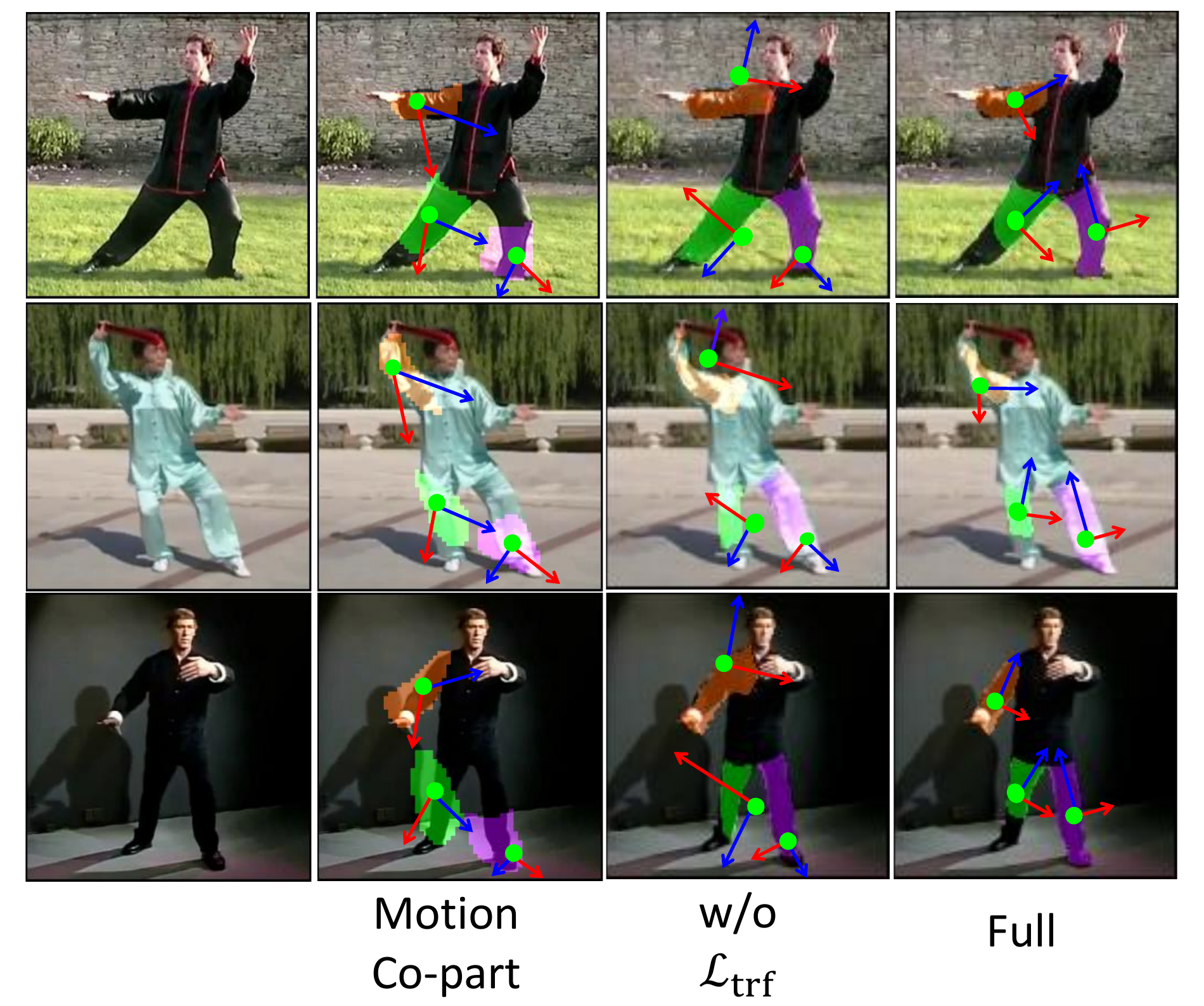}
    \vspace{-2mm}
    \caption{Visual comparison for the interpretability of intermediate affine transformations, which are generated using our method with full loss, disable ${\cal L}_{\text{trf}}$ and method in Motion Co-part respectively.   }
    \label{fig:rotation_vis}
    \vspace{-6mm}
\end{figure}

As illustrated in Figure~\ref{fig:rotation_vis}, $\mathcal{L}_\text{rots}$ and $\mathcal{L}_\text{tran}$ play a critical role in bringing the estimated part transformation with explainable meaning. The part transformations learned without these losses only loosely correlate to the global pose of the parts, while the transformations estimated using our model align with the motion of the object parts.
\llb{
We find these loss terms very effective when the transformation of each part can be clearly defined, such as in the case of human limbs in the Tai-Chi-HD dataset.
Nonetheless, these loss terms take only marginal effect on faces in the VoxCeleb dataset, though we kept this term as a part of a uniform training process.
}


Package pdftex.def Error: File `figures/ablation.pdf' not found: using draft 
setting.




\section{\llb{Discussion}}
In this paper, we have proposed an unsupervised Co-part segmentation approach, which leverages shape correlation information between different frames in the video to achieve semantic part segmentation. We have designed a novel network structure which achieves self-supervision through a dual procedure of part-assembly to form a closed loop with part-segmentation. 
Additionally, we have developed several new loss functions 
that ensure consistent, compact and meaningful part segmentation and the intermediate transformations with clear explainable physical meaning. We have demonstrated the advantages of our method through a host of studies. 

\begin{figure}[t]
    \centering
   \newcommand{\formattedgraphics}[1]{\includegraphics[width=0.47\columnwidth]{#1}}
    \formattedgraphics{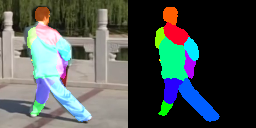}
    \formattedgraphics{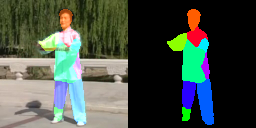}\\%
    \formattedgraphics{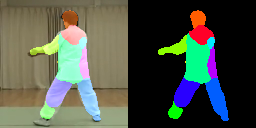}
    \formattedgraphics{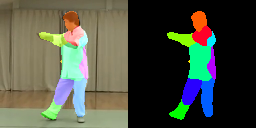}%
    \caption{Failure case. Due to the lack of temporal information, our method can fail in inferring occluded parts and may incorrectly label the limbs when a person turn around. }
    \label{fig:failure_case}
    \vspace{-4mm}
\end{figure}

\llb{
We empirically choose weights in our training to balance the magnitude of each loss term in a preliminary training, except for the image reconstruction loss, which is an order of magnitude larger than the other regularization terms due to its critical role in the training. The performance of our model is not sensitive to the specific choice of these loss weights, and similar video categories can share the loss weights.}
\llb{In order to make a fair comparison, we deliberately use different $K$ in some of the experiments to ensure consistency with the baseline methods, but the performance of the method is not sensitive to the specific value of a large enough $K$. For example, we can achieve correct segmentation using an empirical value of $K = 15$ in all the test cases discussed in the paper. 
}

Several lines are open for future research. 
\llb{First, As showed in Figure~\ref{fig:failure_case}, due to the lack of temporal information, our method can fail in inferring occluded parts and may incorrectly label the limbs when a person turns around. It would be a valuable extension of the current framework to train with image sequences or videos to address such inconsistency issues using the time coherence information embedded in the video.}
\llb{%
Second, although our method allows moderate background motions as exhibited in Tai-Chi-HD and VoxCeleb datasets during training, dramatic background change can interfere the training stage and degrade the performance. Extending our method to support training on videos with dramatic background change is a viable future work.%
}
Another interesting direction would be additionally identifying joint positions, which would significantly support a more diverse range of applications. Lastly, extending the study of co-part segmentation to 3D makes another meaningful future work.

\section*{Acknowledgements}
\llb{
We thank the anonymous reviewers for their constructive comments. 
This work was supported in part by National Key R$\&$D Program of China (2020AAA0105200, 2019YFF0302900) and Beijing Academy of Artificial Intelligence (BAAI).%
}


\clearpage

\bibliography{egbib}

\begin{thebibliography}{34}
\providecommand{\natexlab}[1]{#1}
\providecommand{\url}[1]{\texttt{#1}}
\expandafter\ifx\csname urlstyle\endcsname\relax
  \providecommand{\doi}[1]{doi: #1}\else
  \providecommand{\doi}{doi: \begingroup \urlstyle{rm}\Url}\fi

\bibitem[Bulat \& Tzimiropoulos(2017)Bulat and Tzimiropoulos]{bulat2017far}
Bulat, A. and Tzimiropoulos, G.
\newblock How far are we from solving the 2d \& 3d face alignment problem?(and
  a dataset of 230,000 3d facial landmarks).
\newblock In \emph{Proceedings of the IEEE International Conference on Computer
  Vision}, pp.\  1021--1030, 2017.

\bibitem[Cao et~al.(2019)Cao, Hidalgo, Simon, Wei, and Sheikh]{cao2019openpose}
Cao, Z., Hidalgo, G., Simon, T., Wei, S.-E., and Sheikh, Y.
\newblock Openpose: realtime multi-person 2d pose estimation using part
  affinity fields.
\newblock \emph{IEEE Transactions on Pattern Analysis and Machine
  Intelligence}, 43\penalty0 (1):\penalty0 172--186, 2019.

\bibitem[{Chang} \& {Demiris}(2018){Chang} and {Demiris}]{Chang18}
{Chang}, H.~J. and {Demiris}, Y.
\newblock Highly articulated kinematic structure estimation combining motion
  and skeleton information.
\newblock \emph{IEEE Transactions on Pattern Analysis and Machine
  Intelligence}, 40\penalty0 (9):\penalty0 2165--2179, 2018.

\bibitem[Collins et~al.(2018)Collins, Achanta, and Susstrunk]{collins2018deep}
Collins, E., Achanta, R., and Susstrunk, S.
\newblock Deep feature factorization for concept discovery.
\newblock In \emph{Proceedings of the European Conference on Computer Vision},
  pp.\  336--352, 2018.

\bibitem[Eigen \& Fergus(2015)Eigen and Fergus]{eigen_predicting_2015}
Eigen, D. and Fergus, R.
\newblock Predicting depth, surface normals and semantic labels with a common
  multi-scale convolutional architecture.
\newblock In \emph{Proceedings of the IEEE International Conference on Computer
  Vision}, pp.\  2650--2658, 2015.

\bibitem[Eslami \& Williams(2012)Eslami and Williams]{eslami2012generative}
Eslami, S. and Williams, C.
\newblock A generative model for parts-based object segmentation.
\newblock \emph{Advances in Neural Information Processing Systems},
  25:\penalty0 100--107, 2012.

\bibitem[Greff et~al.(2017)Greff, van Steenkiste, and
  Schmidhuber]{NIPS2017_nem}
Greff, K., van Steenkiste, S., and Schmidhuber, J.
\newblock Neural expectation maximization.
\newblock In \emph{Advances in Neural Information Processing Systems},
  volume~30, pp.\  6691--6701, 2017.

\bibitem[G{\"u}ler et~al.(2018)G{\"u}ler, Neverova, and
  Kokkinos]{guler2018densepose}
G{\"u}ler, R.~A., Neverova, N., and Kokkinos, I.
\newblock Densepose: Dense human pose estimation in the wild.
\newblock In \emph{Proceedings of the IEEE Conference on Computer Vision and
  Pattern Recognition}, pp.\  7297--7306, 2018.

\bibitem[Hung et~al.(2019)Hung, Jampani, Liu, Molchanov, Yang, and
  Kautz]{hung2019scops}
Hung, W.-C., Jampani, V., Liu, S., Molchanov, P., Yang, M.-H., and Kautz, J.
\newblock Scops: Self-supervised co-part segmentation.
\newblock In \emph{Proceedings of the IEEE Conference on Computer Vision and
  Pattern Recognition}, pp.\  869--878, 2019.

\bibitem[Jakab et~al.(2018)Jakab, Gupta, Bilen, and
  Vedaldi]{jakab_unsupervised_2018}
Jakab, T., Gupta, A., Bilen, H., and Vedaldi, A.
\newblock Unsupervised {Learning} of {Object} {Landmarks} through {Conditional}
  {Image} {Generation}.
\newblock \emph{Advances in Neural Information Processing Systems},
  31:\penalty0 4016--4027, 2018.

\bibitem[Johnson et~al.(2016)Johnson, Alahi, and Fei-Fei]{johnson2016vggloss}
Johnson, J., Alahi, A., and Fei-Fei, L.
\newblock Perceptual losses for real-time style transfer and super-resolution.
\newblock In \emph{Proceedings of the European Conference on Computer Vision},
  pp.\  694--711, 2016.

\bibitem[Kanazawa et~al.(2018)Kanazawa, Black, Jacobs, and
  Malik]{kanazawa2018end}
Kanazawa, A., Black, M.~J., Jacobs, D.~W., and Malik, J.
\newblock End-to-end recovery of human shape and pose.
\newblock In \emph{Proceedings of the IEEE Conference on Computer Vision and
  Pattern Recognition}, pp.\  7122--7131, 2018.

\bibitem[Keuper et~al.(2015)Keuper, Andres, and Brox]{Keuper15}
Keuper, M., Andres, B., and Brox, T.
\newblock Motion trajectory segmentation via minimum cost multicuts.
\newblock In \emph{Proceedings of the IEEE International Conference on Computer
  Vision}, December 2015.

\bibitem[Khan et~al.(2015)Khan, Mauro, and Leonardi]{khan2015multi}
Khan, K., Mauro, M., and Leonardi, R.
\newblock Multi-class semantic segmentation of faces.
\newblock In \emph{Proceedings of the IEEE International Conference on Image
  Processing}, pp.\  827--831, 2015.

\bibitem[Lathuili{\`e}re et~al.(2020)Lathuili{\`e}re, Tulyakov, Ricci, Sebe,
  et~al.]{lathuiliere2020motion}
Lathuili{\`e}re, S., Tulyakov, S., Ricci, E., Sebe, N., et~al.
\newblock Motion-supervised co-part segmentation.
\newblock \emph{arXiv preprint arXiv:2004.03234}, 2020.

\bibitem[Lee \& Seung(1999)Lee and Seung]{lee1999learning}
Lee, D.~D. and Seung, H.~S.
\newblock Learning the parts of objects by non-negative matrix factorization.
\newblock \emph{Nature}, 401\penalty0 (6755):\penalty0 788--791, 1999.

\bibitem[Nagrani et~al.(2017)Nagrani, Chung, and
  Zisserman]{nagrani2017voxceleb}
Nagrani, A., Chung, J.~S., and Zisserman, A.
\newblock Voxceleb: a large-scale speaker identification dataset.
\newblock 2017.

\bibitem[Nguyen et~al.(2013)Nguyen, Tran, Phung, and
  Venkatesh]{nguyen2013learning}
Nguyen, T.~D., Tran, T., Phung, D., and Venkatesh, S.
\newblock Learning parts-based representations with nonnegative restricted
  boltzmann machine.
\newblock In \emph{Asian Conference on Machine Learning}, pp.\  133--148, 2013.

\bibitem[{Ochs} et~al.(2014){Ochs}, {Malik}, and {Brox}]{Ochs14}
{Ochs}, P., {Malik}, J., and {Brox}, T.
\newblock Segmentation of moving objects by long term video analysis.
\newblock \emph{IEEE Transactions on Pattern Analysis and Machine
  Intelligence}, 36\penalty0 (6):\penalty0 1187--1200, 2014.

\bibitem[Ross \& Zemel(2006)Ross and Zemel]{ross2006learning}
Ross, D.~A. and Zemel, R.~S.
\newblock Learning parts-based representations of data.
\newblock \emph{Journal of Machine Learning Research}, 7\penalty0 (11), 2006.

\bibitem[Ross et~al.(2010)Ross, Tarlow, and Zemel]{Ross10}
Ross, D.~A., Tarlow, D., and Zemel, R.~S.
\newblock Learning articulated structure and motion.
\newblock \emph{International Journal of Computer Vision}, 88\penalty0
  (2):\penalty0 214--237, 2010.

\bibitem[Sabour et~al.(2020)Sabour, Tagliasacchi, Yazdani, Hinton, and
  Fleet]{sabour2020unsupervised}
Sabour, S., Tagliasacchi, A., Yazdani, S., Hinton, G.~E., and Fleet, D.~J.
\newblock Unsupervised part representation by flow capsules.
\newblock \emph{arXiv preprint arXiv:2011.13920}, 2020.

\bibitem[Siarohin et~al.(2019)Siarohin, Lathuilière, Tulyakov, Ricci, and
  Sebe]{siarohin2020first}
Siarohin, A., Lathuilière, S., Tulyakov, S., Ricci, E., and Sebe, N.
\newblock First order motion model for image animation.
\newblock December 2019.

\bibitem[Siarohin et~al.(2020)Siarohin, Roy, Lathuili{\`e}re, Tulyakov, Ricci,
  and Sebe]{siarohin2020motion}
Siarohin, A., Roy, S., Lathuili{\`e}re, S., Tulyakov, S., Ricci, E., and Sebe,
  N.
\newblock Motion-supervised co-part segmentation.
\newblock \emph{arXiv e-prints}, pp.\  arXiv--2004, 2020.

\bibitem[Simonyan \& Zisserman(2015)Simonyan and Zisserman]{simonyan2015vgg}
Simonyan, K. and Zisserman, A.
\newblock Very deep convolutional networks for large-scale image recognition.
\newblock In \emph{Proceedings of the International Conference on Learning
  Representations}, 2015.

\bibitem[Steenkiste et~al.(2018)Steenkiste, Chang, Greff, and
  Schmidhuber]{steenkiste2018relational_RNEM}
Steenkiste, v.~S., Chang, M., Greff, K., and Schmidhuber, J.
\newblock Relational neural expectation maximization: Unsupervised discovery of
  objects and their interactions.
\newblock \emph{Proceedings of the International Conference on Learning
  Representations}, 2018.

\bibitem[Sturm et~al.(2011)Sturm, Stachniss, and Burgard]{Sturm11}
Sturm, J., Stachniss, C., and Burgard, W.
\newblock A probabilistic framework for learning kinematic models of
  articulated objects.
\newblock \emph{Journal of Artificial Intelligence Research}, 41:\penalty0
  477--526, 2011.

\bibitem[Sun \& Savarese(2011)Sun and Savarese]{sun2011articulated}
Sun, M. and Savarese, S.
\newblock Articulated part-based model for joint object detection and pose
  estimation.
\newblock In \emph{Proceedings of the International Conference on Computer
  Vision}, pp.\  723--730, 2011.

\bibitem[Wang \& Yuille(2015)Wang and Yuille]{wang2015semantic}
Wang, J. and Yuille, A.~L.
\newblock Semantic part segmentation using compositional model combining shape
  and appearance.
\newblock In \emph{Proceedings of the IEEE Conference on Computer Vision and
  Pattern Recognition}, pp.\  1788--1797, 2015.

\bibitem[Xu et~al.(2019)Xu, Liu, Sun, Murphy, Freeman, Tenenbaum, and Wu]{psd}
Xu, Z., Liu, Z., Sun, C., Murphy, K., Freeman, W.~T., Tenenbaum, J.~B., and Wu,
  J.
\newblock Unsupervised discovery of parts, structure, and dynamics.
\newblock In \emph{Proceedings of the International Conference on Learning
  Representations}, 2019.

\bibitem[Xue et~al.(2016)Xue, Wu, Bouman, and Freeman]{xue2016visual}
Xue, T., Wu, J., Bouman, K.~L., and Freeman, W.~T.
\newblock Visual dynamics: Probabilistic future frame synthesis via cross
  convolutional networks.
\newblock 2016.

\bibitem[{Yan} \& {Pollefeys}(2008){Yan} and {Pollefeys}]{Yan08}
{Yan}, J. and {Pollefeys}, M.
\newblock A factorization-based approach for articulated nonrigid shape, motion
  and kinematic chain recovery from video.
\newblock \emph{IEEE Transactions on Pattern Analysis and Machine
  Intelligence}, 30\penalty0 (5):\penalty0 865--877, 2008.

\bibitem[Zhang et~al.(2018)Zhang, Guo, Jin, Luo, He, and
  Lee]{zhang_unsupervised_2018}
Zhang, Y., Guo, Y., Jin, Y., Luo, Y., He, Z., and Lee, H.
\newblock Unsupervised discovery of object landmarks as structural
  representations.
\newblock In \emph{Proceedings of the IEEE Conference on Computer Vision and
  Pattern Recognition}, pp.\  2694--2703, June 2018.

\bibitem[Zhou et~al.(2019)Zhou, Barnes, Jingwan, Jimei, and
  Hao]{zhou_continuity_2019}
Zhou, Y., Barnes, C., Jingwan, L., Jimei, Y., and Hao, L.
\newblock On the continuity of rotation representations in neural networks.
\newblock In \emph{Proceedings of the IEEE Conference on Computer Vision and
  Pattern Recognition}, June 2019.

\end{thebibliography}
\bibliographystyle{icml2021}

\end{document}


\date{}
	\title{ Unsupervised Co-part Segmentation through Assembly\\
		(\emph{Supplementary Material})}
	\author{Qingzhe Gao, Bin Wang, Libin Liu, Baoquan Chen}
	\maketitle

	\section{Outline}
	This document contains implementation details of our method. The content is organized as follows.
	\begin{itemize}
		\itemsep-0.21em
		\item \textbf{Section \ref{sec:structure}:} Network structure.
		\item \textbf{Section \ref{sec:Trainingprogress}:} Implementation details.
	\end{itemize}

	\section{Network Structure} \label{sec:structure}
	
\begin{figure*}[htp]
		\centering
		\includegraphics[scale=0.48]{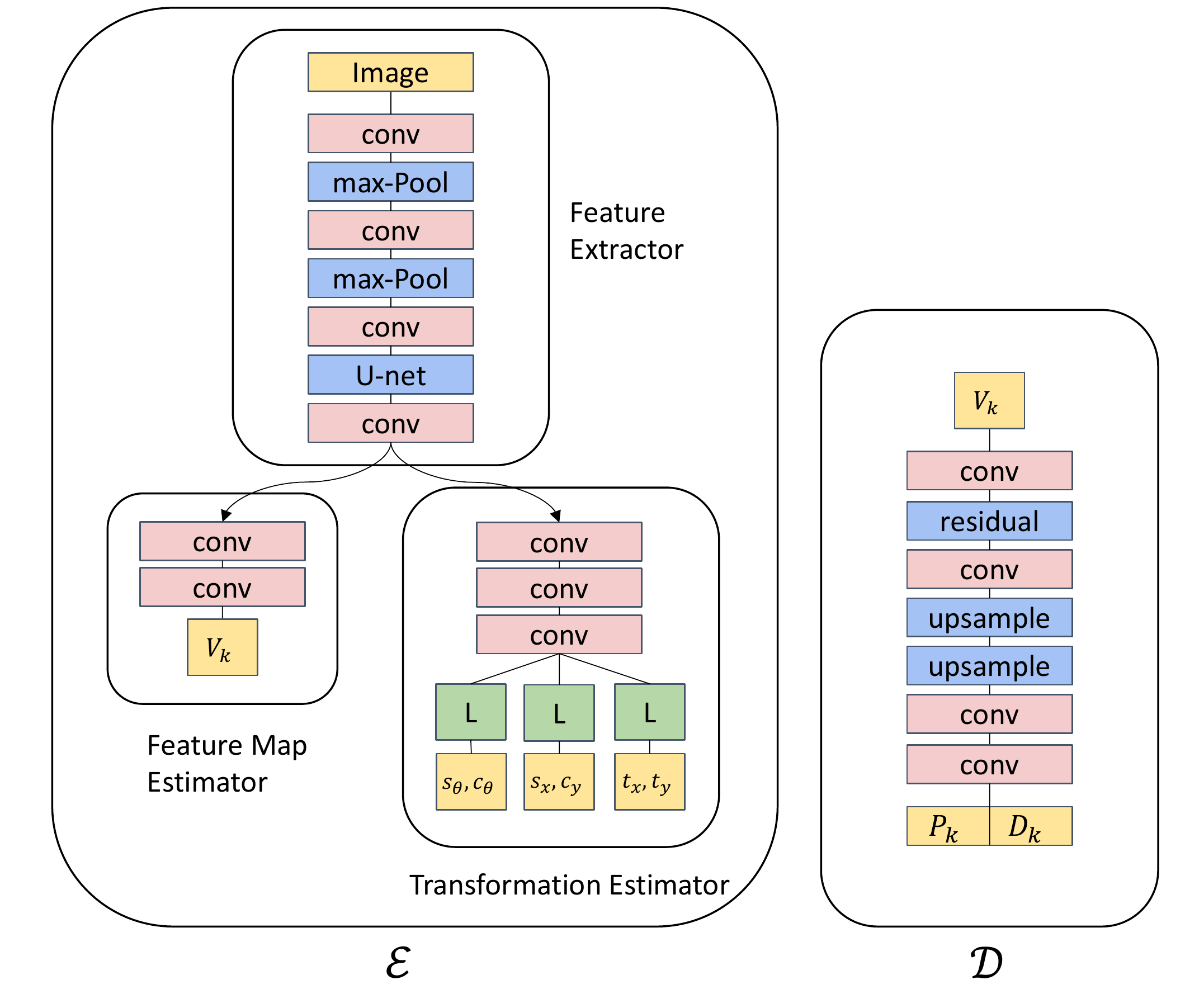}
		\vspace{-2mm}
		\caption{Two major modules of our network: image encoder $\mathcal{E}$ and segment decoder $\mathcal{D}$. The image encoder  $\mathcal{E}$ is composes of three parts: feature extractor, feature map estimator and transformation estimator.}
		\label{figure:network}
	\end{figure*}

	As stated in the main paper, our segmentation network consists of two major modules: image encoder $\mathcal{E}$ and segment decoder $\mathcal{D}$. Their structures are shown in Figure~\ref{figure:network}. 
	In this document, we assume the shape of a tensor is in the format of $C\times H \times W$, where $C$ is the number of channels, $H$ and $W$ are the height and the width of the tensor, respectively.

	\subsection{Image Encoder}
	The image encoder $\mathcal{E}$ includes three parts: feature extractor, feature map estimator, and transformation estimator.

	\textbf{Feature Extractor.}  
	The feature extractor takes an image $I$ of size $(3 \times 128 \times 128)$ as input and computes a feature image $f$ of size $((128\times (K+1)) \times 32\times 32)$, where $(K+1)$ corresponds to $K$ foreground parts and one background part.
	
	The network is constructed based on the standard U-Net architecture~\cite{ronneberger2015unet}, where three cascaded downsampling blocks and the corresponding upsampling blocks are employed. The initial feature count is $512$ and the max feature count is set to $1024$ in our implementation.
	Before being input into the U-Net module, an input image will first pass through three convolutional layers, each followed by a ELU layer and an instance normalization layer, and two max-pooling layers. 
	The output of the U-Net will pass through an additional convolutional layer to compute the final feature image.
	The parameters of these layers can be found in Table~\ref{table:Feature}.
	
	%
	%
	%
    
    \begin{table*}[t]
\caption{Structure of Feature Extractor. Specifically, \emph{conv}, \emph{inst}, \emph{elu} represent the convolution layer, instance normalization layer,and ELU activation respectively.}
\label{table:Feature}
\begin{center}
\begin{small}
\begin{tabular}{ccccccc}
\toprule

Parameter & Kernel&Stride &  Channel in& Channel out &InpRes &OutRes \\

\midrule
conv1+elu+inst & $3 \times 3$&1& 3&64 & $128\times128$ & $128\times128$  \\
maxpool        & $2 \times 2$&1& 64&64 & $128\times128$ & $64\times 64$  \\
conv2+elu+inst & $3 \times 3$&1& 64&256 & $64\times64$ & $64\times64$  \\
maxpool        & $2 \times 2$&1& 256&256 & $64\times 64$ & $32\times 32$  \\
conv3+elu+inst & $3 \times 3$&1& 256&512 & $32\times32$ & $32\times32$  \\
U-Net      & $3 \times 3$&1& 512&1024 & $32\times32$ & $32\times32$  \\
conv4         & $ 3 \times 3$&1& 1024&$128\times (K+1)$ & $32\times32$ & $32\times32$  \\
\bottomrule
\end{tabular}
\end{small}
\end{center}
\end{table*}

    \textbf{Feature Map Estimator}. 
    Based on the output feature image $f$ of the feature extractor, the feature map estimator computes $(K+1)$ feature maps $\{V_k\}, k\in\{0,\dots{},K\}$, each of size $10\times32\times32$.
    We employ two convolutional layers in this module, where a leaky ReLU with $0.2$ negative slope is used as the activation function of the first convolutional layer.
    The parameters of these layers are shown in Table~\ref{table:partmap}.
    \begin{table*}[t]
\caption{Structure of Feature Map Estimator. \emph{lrelu} represents leaky ReLU activation.}
\label{table:partmap}
\begin{center}
\begin{small}
\begin{tabular}{ccccccc}
\toprule

Name & Kernel&Stride &  Channel in&Channel out &InpRes &OutRes \\

\midrule
conv5+lrelu & $ 7 \times 7$&1& $128\times (K+1)$&$32\times (K+1)$ & $32\times32$ & $32\times32$  \\
conv6 & $ 7 \times 7 $&1& $32\times (K+1)$&$10\times (K+1)$ & $32\times32$ & $32\times32$  \\

\bottomrule
\end{tabular}
\end{small}
\end{center}
\end{table*}
    
    \textbf{Transformation Estimator}. 
    Similar to the feature map estimator, the transformation estimator also takes the feature image $f$ as input and computes the transformations of every part. 
    The network first convert $f$ into a one-dimensional feature vector using three convolutional layers, then three separate MLPs are employed to estimate the scaling $(s_x, s_y)$, rotation $(s_{\theta},  c_{\theta})$, and translation $(t_x, t_y)$ of the transformation.
    %
    Note that we only estimate transformations of the $K$ foreground parts. The transformations of the background part is assumed to be identity.
    Table~\ref{table:transformation} reports the parameters of these layers.
    
    
    
    \begin{table*}[t]
\caption{Structure of Transformation Estimator. \emph{linear} represents linear layer.}
\label{table:transformation}
\begin{center}
\begin{small}
\begin{tabular}{ccccccc}
\toprule

Name & Kernel&Stride &  Channel in& Channel out &InpRes &OutRes \\

\midrule
conv7+lrelu & $ 4 \times 4$&2& $128\times (K +1)$&$32\times (K +1)$ & $32\times32$ & $16\times16$  \\
conv8+lrelu & $ 4 \times 4 $&2& $32\times (K +1)$&$16\times (K +1)$ & $16\times16$ & $8\times8$  \\
conv9+lrelu & $ 4 \times 4 $&2& $16\times (K +1)$&$4\times (K +1)$ & $8\times8$ & $4\times4$  \\
resize &-&-& $4\times (K +1)$&$64\times(K +1)$ & $4\times4$ & 1  \\
linear1 &-&-& $64\times(K +1)$&$2\times K$ & 1 & 1  \\
linear2 &-&-& $64\times(K +1)$&$2\times K$ & 1 & 1  \\
linear3 &-&-& $64\times(K +1)$&$2\times K$ & 1 & 1  \\
\bottomrule
\end{tabular}
\end{small}
\end{center}
\end{table*}

    \subsection{Segment Decoder}
    
    \textbf{Segment Decoder}. The segment decoder $\mathcal{D}$ takes a feature map $V_k$ $(10\times32\times32)$ as input, feeds it  through two convolutional  layers, one residual block~\cite{he2016resnet},  and  two upsampling blocks~\cite{johnson2016vggloss}, then finally outputs a tensor with size  $(4\times128\times128)$. The first three channels of this tensor are considered as the part image $P_k$ $(3\times128\times128)$, while the fourth channel is used as the depth map $D_k$ $(1\times128\times128)$. 
    The details are shown in Table ~\ref{table:decoder}.
    \begin{table*}[t]
\caption{Structure of Decoder. \emph{residual} and \emph{upsample} represent residual block~\cite{he2016resnet} and upsampling block~\cite{johnson2016vggloss}.}
\label{table:decoder}
\begin{center}
\begin{small}
\begin{tabular}{ccccccc}
\toprule

Name & Kernel & Stride& Channel in & Channel out &InpRes &OutRes \\

\midrule
conv10+lrelu & $ 7 \times 7 $&1& $10$ &$64$ & $32\times32$ & $32\times32$  \\
residual & $ 3 \times 3 $&1& $64$&$64$ & $32\times32$ & $32\times32$  \\
conv11+lrelu & $ 3 \times 3 $&1& $64$&$128$ & $32\times32$ & $32\times32$  \\
upsample1 & $ 3 \times 3 $&1& $128$&$64$ & $32\times32$ & $64\times64$  \\
upsample2 & $ 3 \times 3 $&1& $64$&$32$ & $64\times64$ & $128\times128$  \\
conv12+lrelu & $ 3 \times 3 $&1& $32$&$16$ & $128\times128$ & $128\times128$  \\
conv13 & $ 3 \times 3 $&1& $16$&$4$ & $128\times128$ & $128\times128$  \\
\bottomrule
\end{tabular}
\end{small}
\end{center}
\end{table*}
    \begin{table*}[t]
\caption{Parameters setting.}
\label{table:Parameters}
\begin{center}
\begin{small}
\begin{tabular}{ccccccccccc}
\toprule

Parameter & $\lambda_1$&$\lambda_2$ &$\lambda_3$&$\lambda_{s}$ &$\lambda_{t}$&$\lambda_{bg}$&$\lambda_{tran}$&$\lambda_{rots}$&$\lambda_{eq}$ &$\lambda_{con}$ \\

\midrule
Tai-Chi-HD&10 &5 &1 &0.1 &1 &0.5   &1 &0.1  &0.02 &0.35\\
VoxCeleb  &10 &5 &1 &0.1 &1 &0.05 &1 &0.05 &0.02 &0.35\\
Exercise  &10 &5 &1 &0.1 &1 &0.5    &1 &0.1 &0.02 &0.35\\
\bottomrule
\end{tabular}
\end{small}
\end{center}
\end{table*}

    \section{Implementation Details}\label{sec:Trainingprogress}
    
    
    Our system is implemented in the PyTorch framework. We train our networks using Adam~\cite{adam2014}, and the learning rate is set to $0.00005$. 
    We use batch size $6$ for $K=10$ and $4$ for $K=15$. The training is performed on one Titan X GPU with $12$GB memory. 
    We terminate the training when the learning progress stalls or exceeds $500,000$ iterations.
    
    In experiment, we set $K$ same to previous work to compare. In practice, the $K$ is not sensitive to the result when $K$ is big enough. The model can automatically select which channels to segment and set redundant channels to be empty. For hand, quadruped, and robot arm, we set the $K$ is $16$, it works well.
    
    We empirically choose weights to balance the magnitude of each loss term in a preliminary training, except for the image reconstruction loss, which is an order of magnitude larger than the other regularization terms due to its critical role in the training.The performance of our model is not sensitive to the specific choice of these loss weights, and similar video categories can share the loss weights.
    
    Table~\ref{table:Parameters} reports the weights of the training losses as described in Section~3.2 of the main paper. Specifically, the vgg loss in the image reconstruction loss is computed in five different resolutions: $128\times 128$, $64\times 64$, $32\times 32$, $16\times 16$, and $8\times 8$. 
    The weights of the corresponding loss terms are set to  $[1.0/32, 1.0/16, 1.0/8, 1.0/4, 1.0]$ as suggested in~\cite{wang2018pix2pixHD}. 
    We use a smaller weight $\lambda_{con}/10$ for the concentration loss computed on canonical parts $M^*_k$. 
    The random transformations used to compute the equivariance loss are generated uniformly in the scaling range $[0.8,1.05]$, the rotation range $[-180 ^\circ{}, 180 ^\circ{}]$, and the translation range $[-0.2,0.2]$.



	
    We train separate networks for each dataset used in this work. The two images of an image pair are both randomly selected from the same video clip.
    We augment the training data using random rotations (in the range of $[-15^\circ, 15^\circ]$) and color jittering to increase the robustness of the network, where the same augmentation is applied to both the images of an image pair.
    
    
	
\begin{figure*}[t]
\begin{center}

\begin{tabular}{c}

\vspace{-0.6mm}
\includegraphics[width=0.95\textwidth]{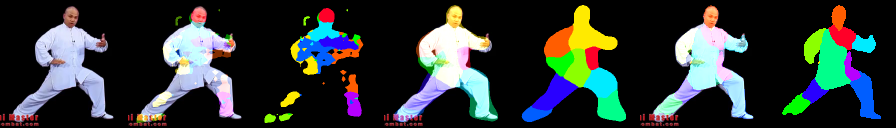}
\\
\vspace{-0.6mm}
\includegraphics[width=0.95\textwidth]{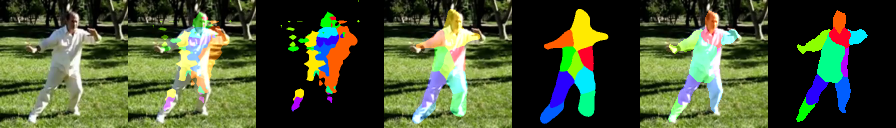}
\\

\vspace{-0.6mm}
\includegraphics[width=0.95\textwidth]{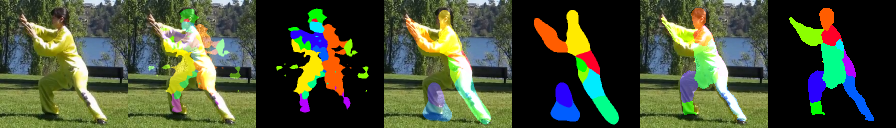}
\\
\vspace{-0.6mm}
\includegraphics[width=0.95\textwidth]{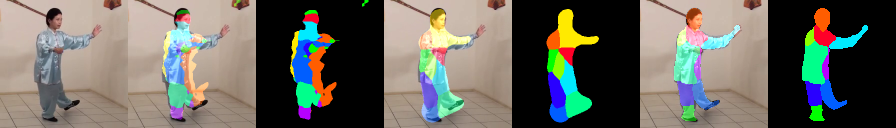}
\\
\vspace{-0.6mm}
\includegraphics[width=0.95\textwidth]{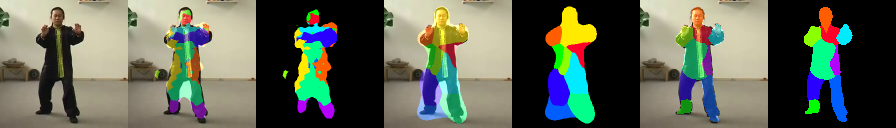}
\\
\vspace{-0.6mm}
\includegraphics[width=0.95\textwidth]{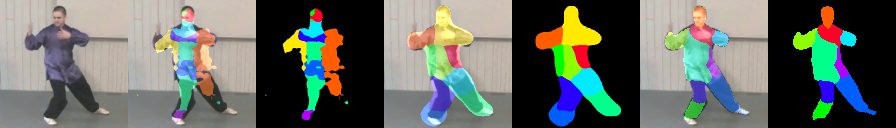}
\\
\vspace{-0.6mm}
\includegraphics[width=0.95\textwidth]{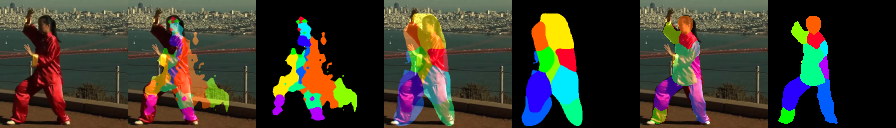}
\\
\vspace{-0.6mm}
\includegraphics[width=0.95\textwidth]{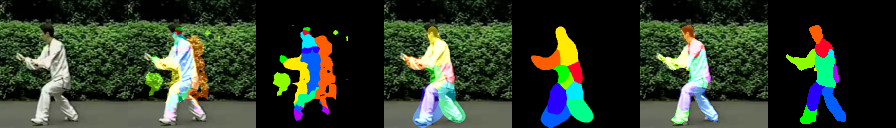}
\\
\vspace{-0.8mm}
\small{Input \qquad\qquad\qquad\qquad 
SCOPS 
\qquad\qquad\qquad\qquad 
Motion Co-part 
\qquad\qquad\qquad\qquad\qquad \qquad
Ours
\qquad
}  
\\
\\
\end{tabular}

\end{center}
\vspace{-6mm}
\caption{Additional results and comparisons on Tai-Chi-HD
} 
\label{figure:vis_taichi1}
\end{figure*}

\begin{figure*}[t]
\begin{center}

\begin{tabular}{c}

\vspace{-0.6mm}
\includegraphics[width=0.95\textwidth]{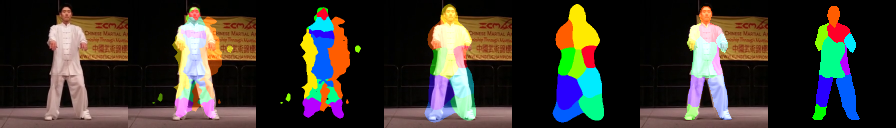}
\\
\vspace{-0.6mm}
\includegraphics[width=0.95\textwidth]{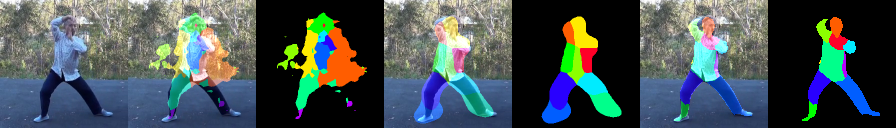}
\\

\vspace{-0.6mm}
\includegraphics[width=0.95\textwidth]{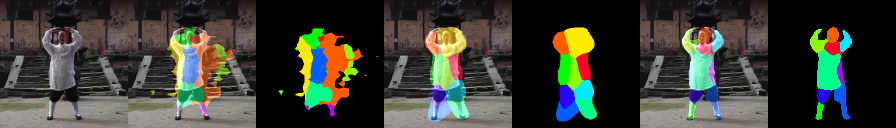}
\\
\vspace{-0.6mm}
\includegraphics[width=0.95\textwidth]{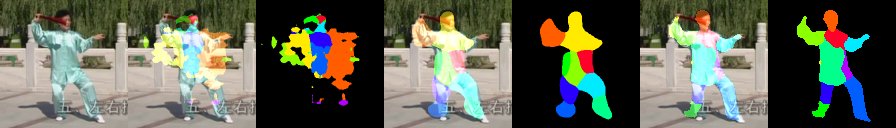}
\\
\vspace{-0.6mm}
\includegraphics[width=0.95\textwidth]{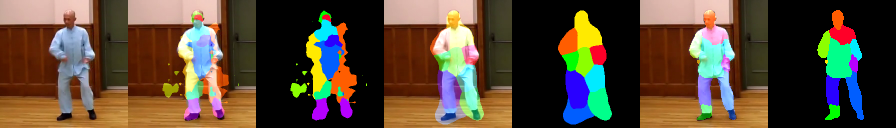}
\\
\vspace{-0.6mm}
\includegraphics[width=0.95\textwidth]{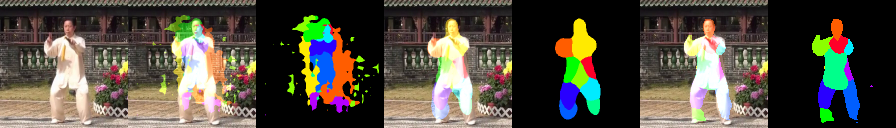}
\\
\vspace{-0.6mm}
\includegraphics[width=0.95\textwidth]{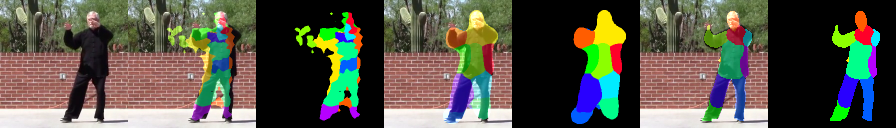}
\\
\vspace{-0.6mm}
\includegraphics[width=0.95\textwidth]{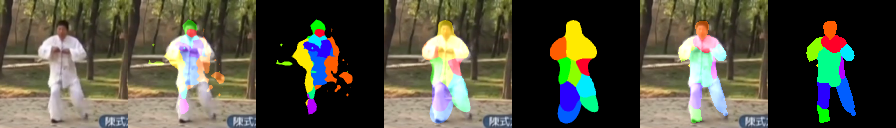}
\\
\vspace{-0.8mm}
\small{Input \qquad\qquad\qquad\qquad 
SCOPS 
\qquad\qquad\qquad\qquad 
Motion Co-part 
\qquad\qquad\qquad\qquad\qquad \qquad
Ours
\qquad
}  
\\
\\
\end{tabular}

\end{center}
\vspace{-6mm}
\caption{Additional results and comparisons on Tai-Chi-HD (cont.)
} 
\label{figure:vis_taichi2}
\end{figure*}

\begin{figure*}[t]
\begin{center}

\begin{tabular}{c}

\vspace{-0.6mm}
\includegraphics[width=0.95\textwidth]{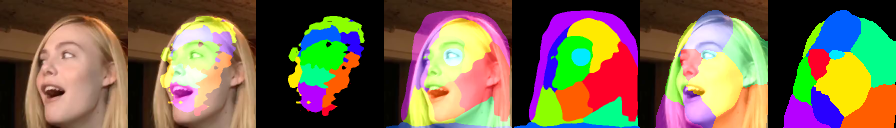}
\\
\vspace{-0.6mm}
\includegraphics[width=0.95\textwidth]{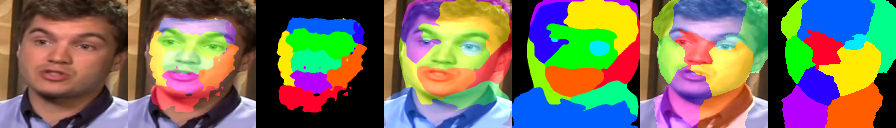}
\\

\vspace{-0.6mm}
\includegraphics[width=0.95\textwidth]{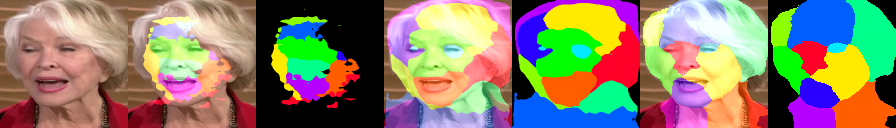}
\\
\vspace{-0.6mm}
\includegraphics[width=0.95\textwidth]{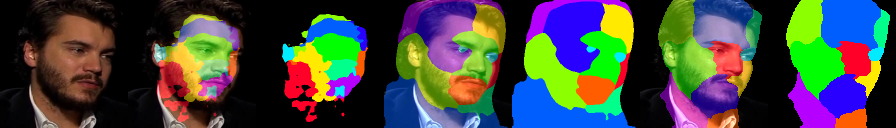}
\\
\vspace{-0.6mm}
\includegraphics[width=0.95\textwidth]{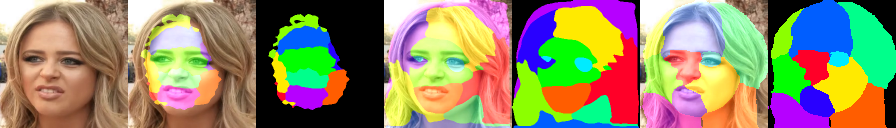}
\\
\vspace{-0.6mm}
\includegraphics[width=0.95\textwidth]{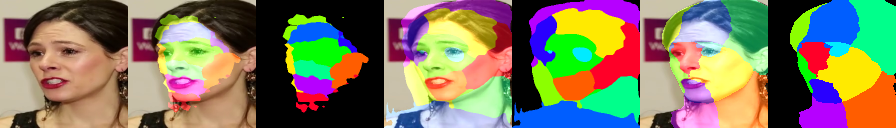}
\\
\vspace{-0.6mm}
\includegraphics[width=0.95\textwidth]{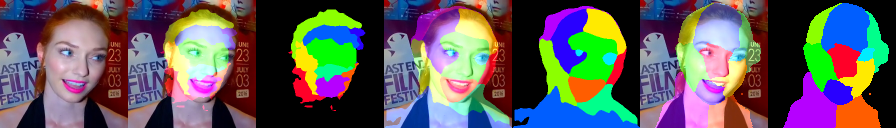}
\\
\vspace{-0.6mm}
\includegraphics[width=0.95\textwidth]{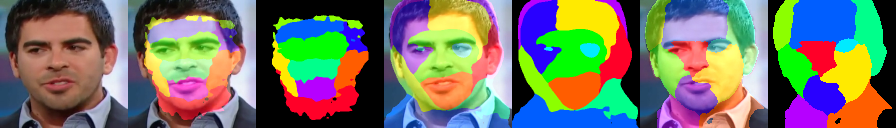}
\\
\vspace{-0.8mm}
\small{Input \qquad\qquad\qquad\qquad 
SCOPS 
\qquad\qquad\qquad\qquad 
Motion Co-part 
\qquad\qquad\qquad\qquad\qquad \qquad
Ours
\qquad
}  
\\
\\
\end{tabular}

\end{center}
\vspace{-6mm}
\caption{Additional results and comparisons on VoxCeleb
} 
\label{figure:vis_vox1}
\end{figure*}

\begin{figure*}[t]
\begin{center}

\begin{tabular}{c}

\vspace{-0.6mm}
\includegraphics[width=0.95\textwidth]{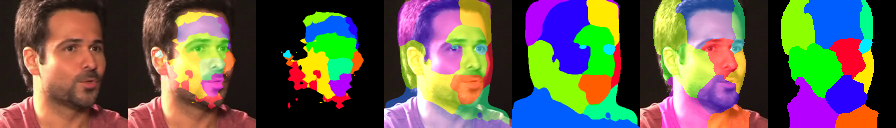}
\\
\vspace{-0.6mm}
\includegraphics[width=0.95\textwidth]{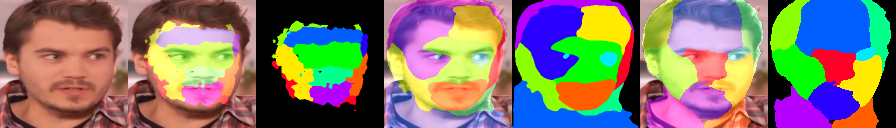}
\\

\vspace{-0.6mm}
\includegraphics[width=0.95\textwidth]{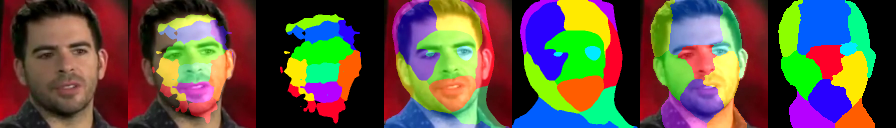}
\\
\vspace{-0.6mm}
\includegraphics[width=0.95\textwidth]{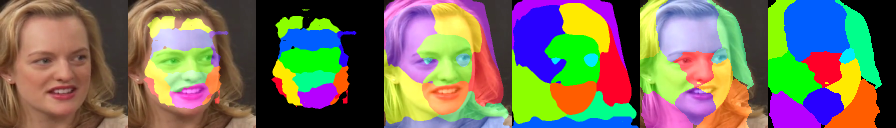}
\\
\vspace{-0.6mm}
\includegraphics[width=0.95\textwidth]{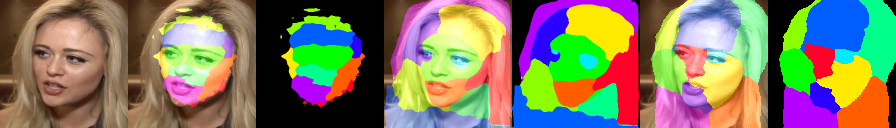}
\\
\vspace{-0.6mm}
\includegraphics[width=0.95\textwidth]{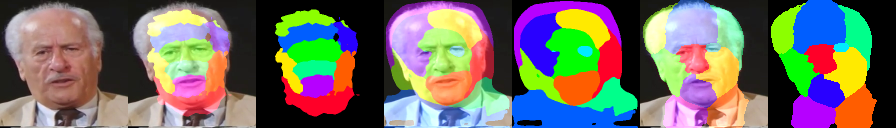}
\\
\vspace{-0.6mm}
\includegraphics[width=0.95\textwidth]{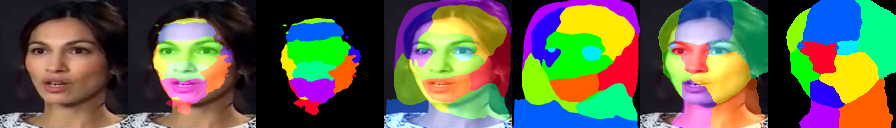}
\\
\vspace{-0.6mm}
\includegraphics[width=0.95\textwidth]{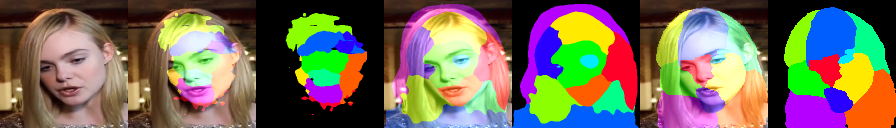}
\\
\vspace{-0.8mm}
\small{Input \qquad\qquad\qquad\qquad 
SCOPS 
\qquad\qquad\qquad\qquad 
Motion Co-part 
\qquad\qquad\qquad\qquad\qquad \qquad
Ours
\qquad
}  
\\
\\
\end{tabular}

\end{center}
\vspace{-6mm}
\caption{Additional results and comparisons on VoxCeleb (cont.)
} 
\label{figure:vis_vox2}
\end{figure*}

\begin{figure*}[t]
\begin{center}

\begin{tabular}{c cc cc cc cc}

\vspace{-0.6mm}
\includegraphics[width=0.1\textwidth]{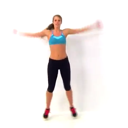}
&\includegraphics[width=0.1\textwidth]{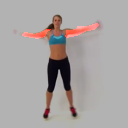}
&\includegraphics[width=0.1\textwidth]{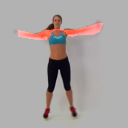}
&\includegraphics[width=0.1\textwidth]{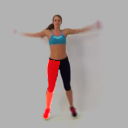}
&\includegraphics[width=0.1\textwidth]{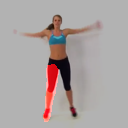}
&\includegraphics[width=0.1\textwidth]{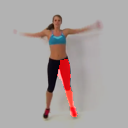}
&\includegraphics[width=0.1\textwidth]{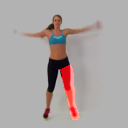}
&\includegraphics[width=0.1\textwidth]{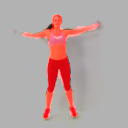}
&\includegraphics[width=0.1\textwidth]{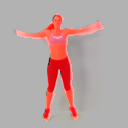}
\\
\includegraphics[width=0.1\textwidth]{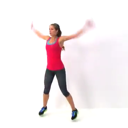}
&\includegraphics[width=0.1\textwidth]{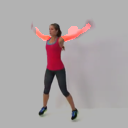}
&\includegraphics[width=0.1\textwidth]{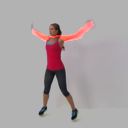}
&\includegraphics[width=0.1\textwidth]{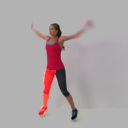}
&\includegraphics[width=0.1\textwidth]{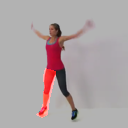}
&\includegraphics[width=0.1\textwidth]{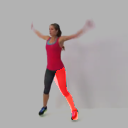}
&\includegraphics[width=0.1\textwidth]{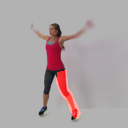}
&\includegraphics[width=0.1\textwidth]{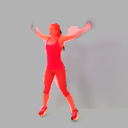}
&\includegraphics[width=0.1\textwidth]{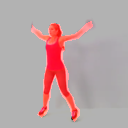}
\\
\includegraphics[width=0.1\textwidth]{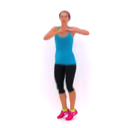}
&\includegraphics[width=0.1\textwidth]{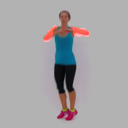}
&\includegraphics[width=0.1\textwidth]{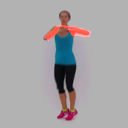}
&\includegraphics[width=0.1\textwidth]{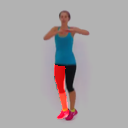}
&\includegraphics[width=0.1\textwidth]{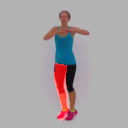}
&\includegraphics[width=0.1\textwidth]{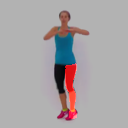}
&\includegraphics[width=0.1\textwidth]{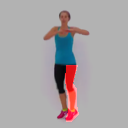}
&\includegraphics[width=0.1\textwidth]{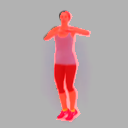}
&\includegraphics[width=0.1\textwidth]{supplementary/exercise/3_full_ground.png}
\\

\includegraphics[width=0.1\textwidth]{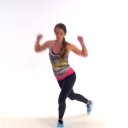}
&\includegraphics[width=0.1\textwidth]{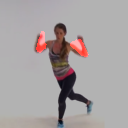}
&\includegraphics[width=0.1\textwidth]{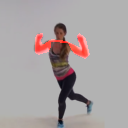}
&\includegraphics[width=0.1\textwidth]{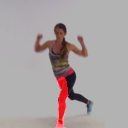}
&\includegraphics[width=0.1\textwidth]{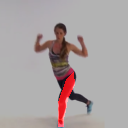}
&\includegraphics[width=0.1\textwidth]{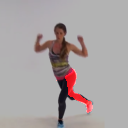}
&\includegraphics[width=0.1\textwidth]{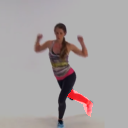}
&\includegraphics[width=0.1\textwidth]{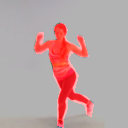}
&\includegraphics[width=0.1\textwidth]{supplementary/exercise/4_full_ground.png}
\\

\includegraphics[width=0.1\textwidth]{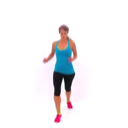}
&\includegraphics[width=0.1\textwidth]{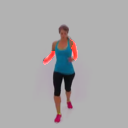}
&\includegraphics[width=0.1\textwidth]{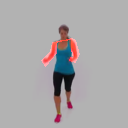}
&\includegraphics[width=0.1\textwidth]{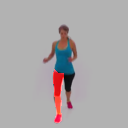}
&\includegraphics[width=0.1\textwidth]{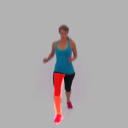}
&\includegraphics[width=0.1\textwidth]{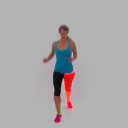}
&\includegraphics[width=0.1\textwidth]{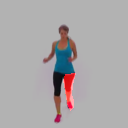}
&\includegraphics[width=0.1\textwidth]{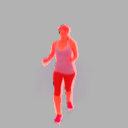}
&\includegraphics[width=0.1\textwidth]{supplementary/exercise/6_full_ground.png}
\\
\includegraphics[width=0.1\textwidth]{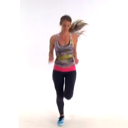}
&\includegraphics[width=0.1\textwidth]{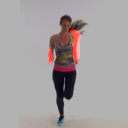}
&\includegraphics[width=0.1\textwidth]{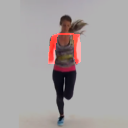}
&\includegraphics[width=0.1\textwidth]{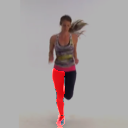}
&\includegraphics[width=0.1\textwidth]{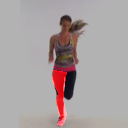}
&\includegraphics[width=0.1\textwidth]{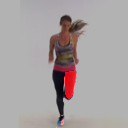}
&\includegraphics[width=0.1\textwidth]{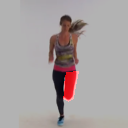}
&\includegraphics[width=0.1\textwidth]{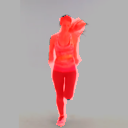}
&\includegraphics[width=0.1\textwidth]{supplementary/exercise/5_full_ground.png}
\\
\includegraphics[width=0.1\textwidth]{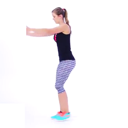}
&\includegraphics[width=0.1\textwidth]{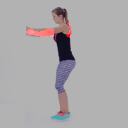}
&\includegraphics[width=0.1\textwidth]{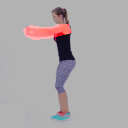}
&\includegraphics[width=0.1\textwidth]{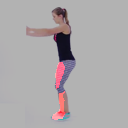}
&\includegraphics[width=0.1\textwidth]{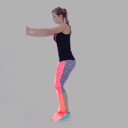}
&\includegraphics[width=0.1\textwidth]{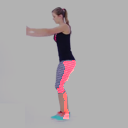}
&\includegraphics[width=0.1\textwidth]{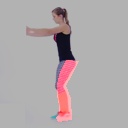}
&\includegraphics[width=0.1\textwidth]{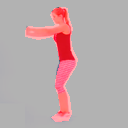}
&\includegraphics[width=0.1\textwidth]{supplementary/exercise/8_full_ground.png}
\\
\includegraphics[width=0.1\textwidth]{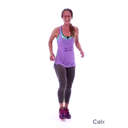}
&\includegraphics[width=0.1\textwidth]{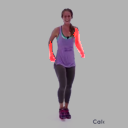}
&\includegraphics[width=0.1\textwidth]{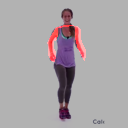}
&\includegraphics[width=0.1\textwidth]{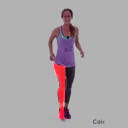}
&\includegraphics[width=0.1\textwidth]{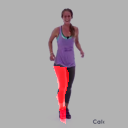}
&\includegraphics[width=0.1\textwidth]{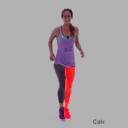}
&\includegraphics[width=0.1\textwidth]{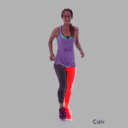}
&\includegraphics[width=0.1\textwidth]{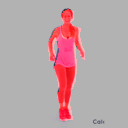}
&\includegraphics[width=0.1\textwidth]{supplementary/exercise/10_full_ground.png}
\\

\includegraphics[width=0.1\textwidth]{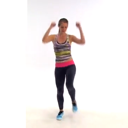}
&\includegraphics[width=0.1\textwidth]{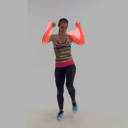}
&\includegraphics[width=0.1\textwidth]{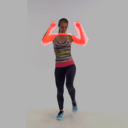}
&\includegraphics[width=0.1\textwidth]{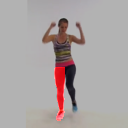}
&\includegraphics[width=0.1\textwidth]{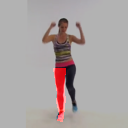}
&\includegraphics[width=0.1\textwidth]{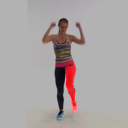}
&\includegraphics[width=0.1\textwidth]{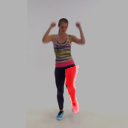}
&\includegraphics[width=0.1\textwidth]{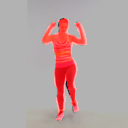}
&\includegraphics[width=0.1\textwidth]{supplementary/exercise/11_full_ground.png}
\\
\vspace{-0.8mm}
\small{Input} 
&\small{Arm} 
& \small{GT} 
& \small{Leg(L)} 
& \small{GT} 
&\small{Leg(R)} 
& \small{GT } 
& \small{Full} 
& \small{GT}
\\
\\
\end{tabular}

\end{center}
\vspace{-6mm}
\caption{Segmentation results on Exercise dataset. We show both our results and the ground-truth (GT) segmentation provided by the dataset. The part masks are superimposed on the input images.
} \label{figure:exercise1}
\end{figure*}

\begin{figure*}[t]
\begin{center}

\begin{tabular}{c cc cc cc cc}

\vspace{-0.6mm}
\includegraphics[width=0.1\textwidth]{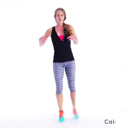}
&\includegraphics[width=0.1\textwidth]{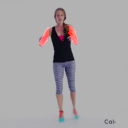}
&\includegraphics[width=0.1\textwidth]{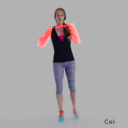}
&\includegraphics[width=0.1\textwidth]{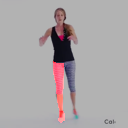}
&\includegraphics[width=0.1\textwidth]{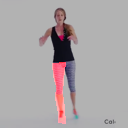}
&\includegraphics[width=0.1\textwidth]{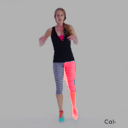}
&\includegraphics[width=0.1\textwidth]{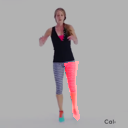}
&\includegraphics[width=0.1\textwidth]{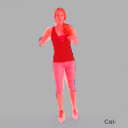}
&\includegraphics[width=0.1\textwidth]{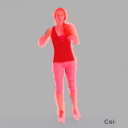}
\\
\includegraphics[width=0.1\textwidth]{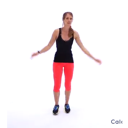}
&\includegraphics[width=0.1\textwidth]{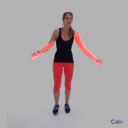}
&\includegraphics[width=0.1\textwidth]{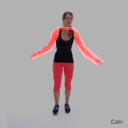}
&\includegraphics[width=0.1\textwidth]{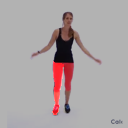}
&\includegraphics[width=0.1\textwidth]{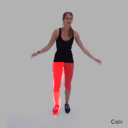}
&\includegraphics[width=0.1\textwidth]{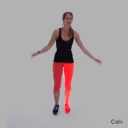}
&\includegraphics[width=0.1\textwidth]{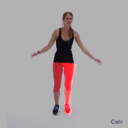}
&\includegraphics[width=0.1\textwidth]{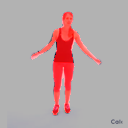}
&\includegraphics[width=0.1\textwidth]{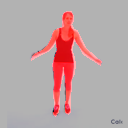}
\\
\includegraphics[width=0.1\textwidth]{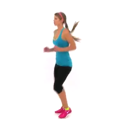}
&\includegraphics[width=0.1\textwidth]{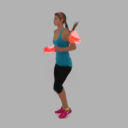}
&\includegraphics[width=0.1\textwidth]{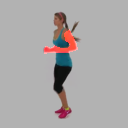}
&\includegraphics[width=0.1\textwidth]{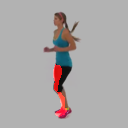}
&\includegraphics[width=0.1\textwidth]{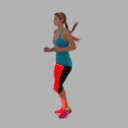}
&\includegraphics[width=0.1\textwidth]{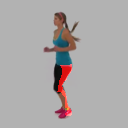}
&\includegraphics[width=0.1\textwidth]{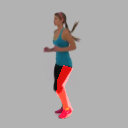}
&\includegraphics[width=0.1\textwidth]{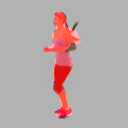}
&\includegraphics[width=0.1\textwidth]{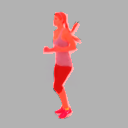}
\\
\includegraphics[width=0.1\textwidth]{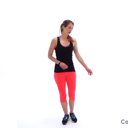}
&\includegraphics[width=0.1\textwidth]{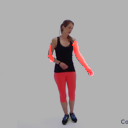}
&\includegraphics[width=0.1\textwidth]{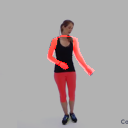}
&\includegraphics[width=0.1\textwidth]{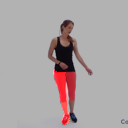}
&\includegraphics[width=0.1\textwidth]{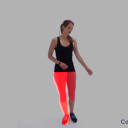}
&\includegraphics[width=0.1\textwidth]{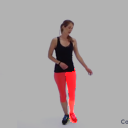}
&\includegraphics[width=0.1\textwidth]{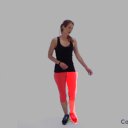}
&\includegraphics[width=0.1\textwidth]{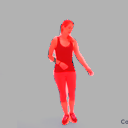}
&\includegraphics[width=0.1\textwidth]{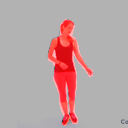}
\\

\includegraphics[width=0.1\textwidth]{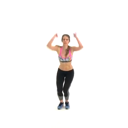}
&\includegraphics[width=0.1\textwidth]{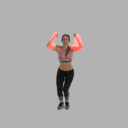}
&\includegraphics[width=0.1\textwidth]{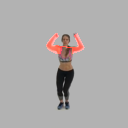}
&\includegraphics[width=0.1\textwidth]{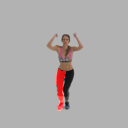}
&\includegraphics[width=0.1\textwidth]{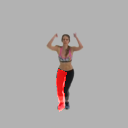}
&\includegraphics[width=0.1\textwidth]{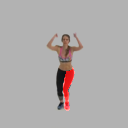}
&\includegraphics[width=0.1\textwidth]{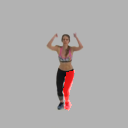}
&\includegraphics[width=0.1\textwidth]{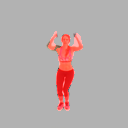}
&\includegraphics[width=0.1\textwidth]{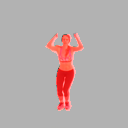}
\\
\includegraphics[width=0.1\textwidth]{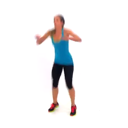}
&\includegraphics[width=0.1\textwidth]{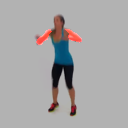}
&\includegraphics[width=0.1\textwidth]{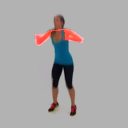}
&\includegraphics[width=0.1\textwidth]{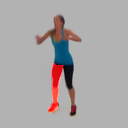}
&\includegraphics[width=0.1\textwidth]{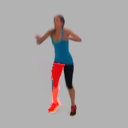}
&\includegraphics[width=0.1\textwidth]{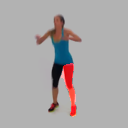}
&\includegraphics[width=0.1\textwidth]{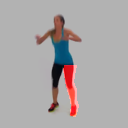}
&\includegraphics[width=0.1\textwidth]{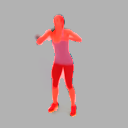}
&\includegraphics[width=0.1\textwidth]{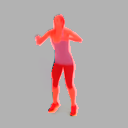}
\\
\includegraphics[width=0.1\textwidth]{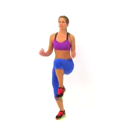}
&\includegraphics[width=0.1\textwidth]{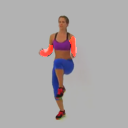}
&\includegraphics[width=0.1\textwidth]{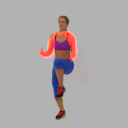}
&\includegraphics[width=0.1\textwidth]{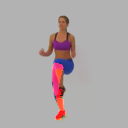}
&\includegraphics[width=0.1\textwidth]{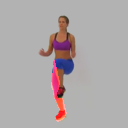}
&\includegraphics[width=0.1\textwidth]{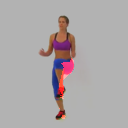}
&\includegraphics[width=0.1\textwidth]{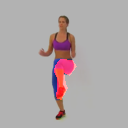}
&\includegraphics[width=0.1\textwidth]{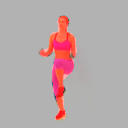}
&\includegraphics[width=0.1\textwidth]{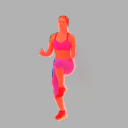}
\\
\includegraphics[width=0.1\textwidth]{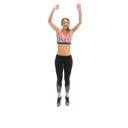}
&\includegraphics[width=0.1\textwidth]{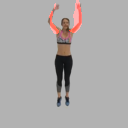}
&\includegraphics[width=0.1\textwidth]{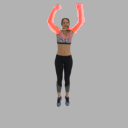}
&\includegraphics[width=0.1\textwidth]{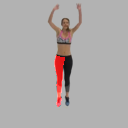}
&\includegraphics[width=0.1\textwidth]{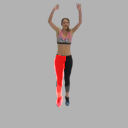}
&\includegraphics[width=0.1\textwidth]{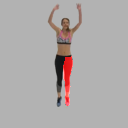}
&\includegraphics[width=0.1\textwidth]{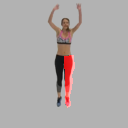}
&\includegraphics[width=0.1\textwidth]{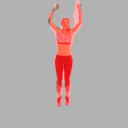}
&\includegraphics[width=0.1\textwidth]{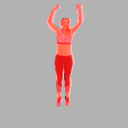}
\\
\includegraphics[width=0.1\textwidth]{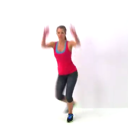}
&\includegraphics[width=0.1\textwidth]{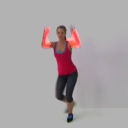}
&\includegraphics[width=0.1\textwidth]{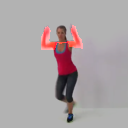}
&\includegraphics[width=0.1\textwidth]{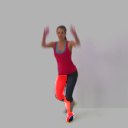}
&\includegraphics[width=0.1\textwidth]{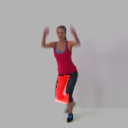}
&\includegraphics[width=0.1\textwidth]{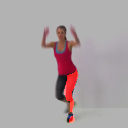}
&\includegraphics[width=0.1\textwidth]{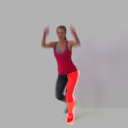}
&\includegraphics[width=0.1\textwidth]{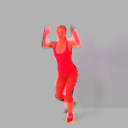}
&\includegraphics[width=0.1\textwidth]{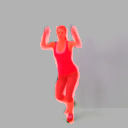}
\\

\vspace{-0.8mm}
\small{Input} 
&\small{Arm} 
& \small{GT} 
& \small{Leg(L)} 
& \small{GT} 
&\small{Leg(R)} 
& \small{GT } 
& \small{Full} 
& \small{GT}
\\
\\
\end{tabular}

\end{center}
\vspace{-6mm}
\caption{
Segmentation results on Exercise dataset (cont.). We show both our results and the ground-truth (GT) segmentation provided by the dataset. The part masks are superimposed on the input images.
} \label{figure:exercise2}
\end{figure*}

	
	We have included some of our results and comparisons to the state-of-the-art approaches in the main paper. For additional results and comparison, please refer to Figure~\ref{figure:vis_taichi1}, Figure~\ref{figure:vis_taichi2}, Figure~\ref{figure:vis_vox1}, Figure~\ref{figure:vis_vox2}, Figure~\ref{figure:exercise1}, and Figure~\ref{figure:exercise2} in this document, as well as the the supplementary video.
	
	\setlength{\tabcolsep}{1pt}
\renewcommand{\arraystretch}{1}

\setlength{\tabcolsep}{1pt}
\renewcommand{\arraystretch}{1}

    \setlength{\tabcolsep}{1pt}
\renewcommand{\arraystretch}{1}

\setlength{\tabcolsep}{1pt}
\renewcommand{\arraystretch}{1}

    \setlength{\tabcolsep}{1pt}
\renewcommand{\arraystretch}{1}

\setlength{\tabcolsep}{1pt}
\renewcommand{\arraystretch}{1}

\clearpage
\newpage{}
\bibliography{egbib}
\bibliographystyle{icml2021}


\date{}
	\title{AONet: Unifying Everything for High-Definition Intrinsics \\
		(\emph{Supplementary Material})}
	
	\author{Paper ID 2572}
	
	\maketitle
	
	\graphicspath{{supplementary_results/images_jpg/network_ablation_study/}{supplementary_results/images_jpg/dataset_comparison/}{supplementary_results/images_jpg/CGIntrinsics_data/}{supplementary_results/images_jpg/IIW_results_comparison/}{supplementary_results/images_jpg/synthetic_smoothing_comparison/}{supplementary_results/images_jpg/IIW_smoothing_comparison/}{supplementary_results/temporary_images_HDI/}{supplementary_results/images_jpg/results_for_albedoEdge/}{supplementary_results/more_IIW_results_1121/}{supplementary_results/images_jpg/selected_multistep_results1122/}{supplementary_results/images_jpg/selected_smoothing_results1122/}{supplementary_results/images_jpg/real_data_results/}{supplementary_results/images_jpg/interior_jpg/}{supplementary_results/analyze_effective_receptive_field/}}

	\section{Outline}
	This document contains more implementation details and experimental results that could not be included in the main paper due to space constraints. Note we demonstrate around 50 examples both on IIW/SAW real data and from unseen data source, in order to show the robustness of our algorithm and its superior performance over existing approaches. The remaining content is organized as follows.
	\begin{itemize}
		\itemsep-0.21em
		\item \textbf{Section \ref{sec:implementation_details}:} Implementation details and network structure.
		\item \textbf{Section \ref{sec:refinement_module}:} Illustration and Justification of our image refinement module.
		\item \textbf{Section \ref{sec:network_analysis}:} Feature analysis of the simple two-network solution.
		\item \textbf{Section \ref{sec:dataset_comparison}:} More examples from our proposed dataset, compared with CGIntrinsics \cite{li2018cgintrinsics}  and InteriorNet \cite{li2018interiornet} dataset.
		\item \textbf{Section \ref{sec:results_iiw_saw}:} More visual comparisons on the IIW/SAW dataset.
		\item \textbf{Section \ref{sec:results_outdoor_scene}:} More visual comparisons on the unseen data source.
	\end{itemize}
	
	\clearpage

	\section{Implementation details and network structure} \label{sec:implementation_details}
	
	In this section, we introduce our implementation details and network structure.
	
	\textbf{Implementation Details.} Our framework is implemented in PyTorch framework with mini-batch size of 8. Our network is optimized by Adam. Its initial learning rate is set as 0.01, and decreased to 0.0005 while finetuning IIW/SAW dataset. The learning rate is maintained to be the same as 0.01 while finetuing MIT dataset, since it is more different from our synthetic training data in all aspects than IIW/SAW dataset.
	
	\textbf{Network Structure.}  The intrinsic estimation network consists of 20 successive convolution layers, of which the first three encode the input image into high dimensional deep features, and the last three decode the intermediate features back to output image. All the other convolution layers are arranged into residual blocks \cite{he2016deep} that have been widely used in the deep learning community. Moreover, we deploy dilated convolution to explicitly increase the receptive field of the neural network. The reason for this design is to meet the specific characteristics of intrinsic images, for example, either piece-wise constancy or shadow removal for albedo require consideration of distant pixels. We also downsample the feature maps before the residual blocks by one fourth and upsample them back to the original resolution after the residual blocks to accelerate the training procedure. All the convolution layers share the same 3$\times$3 kernel size, output 64 channels of feature maps, and are followed by instance normalization (IN) and ReLU layers, except for the last one. 
	
	Its detailed network structure is shown in Table \ref{table:estimation_network}, while its corresponding intrinsic decision network $f_{ID}$  is revealed in Table \ref{table:decision_network}. 
	
	\input{table-estimation-network.tex}
	\input{table-decision-network.tex}
	
	\clearpage

	\section{Illustration and justification of the image refinement module $f_{IR}$} \label{sec:refinement_module}
	
	\subsection{ How is the image refinement module $f_{IR}$ trained and evaluated? }
	In Figure \ref{figure:refinement_module}, we illustrate the difference between the training and testing phase of our proposed image refinement module. In the training stage, we feed our rendered input image ($I$) and an image structure map ($E_R$) calculated from the label ($R^*$), which is \cite{L1smooth}'s processed image of $I$, into the neural network, and generate its output ($R$). In the evaluation stage, we feed the albedo image ($A$) and the albedo structure ($E_A$) predicted from the intrinsic estimation network to the refinement module to generate the final albedo ($A_R$).
	
	As can be seen, the refined image shares the similar image structure with the input structure map, and similar color information with the input image. It learns to refine the input image guided by the input structure map implicitly in the deep neural network, which is dramatically different from any previous guided image filtering process.
	
	\begin{figure*}[htp]
		\centering
		\includegraphics[scale=0.48]{refinement_network.png}
		\vspace{-2mm}
		\caption{Training and evaluation phase of the refinement module $f_{IR}$.}
		\label{figure:refinement_module}
	\end{figure*}
	
	\subsection{Significance of the  image refinement module}
	\cite{fan2018revisiting} integrated a guided domain filter in its pipeline as a post-processed refinement step, which is the closest to ours in spirit. Therefore, in order to justify the effectiveness of our image refinement module, we replace it with the domain filter \cite{fan2018revisiting} and evaluate its performance on the IIW test data, which achieves \textbf{20.90} on the WHDR error metric. It lags behind our \textbf{17.10} by a large margin.
	
	Note their domain filter is supervised by the albedo and jointly trained within the whole framework, while our refinement module doesn't learn any prior of the intrinsic information and is independently trained.
	
	\clearpage

	\section{Feature analysis of two-network solution} \label{sec:network_analysis}
	
	We further analyze the influence of different feature channels to the intrinsic image prediction by training two fully convolution neural networks, which share the same network structure with our intrinsic estimation network, for albedo and shading prediction separately. We illustrate the activeness of different feature channels in Figure \ref{figure:analysis}, similarly to Figure 6 in the main paper.
	
	We can see that the feature distribution of albedo and shading are dramatically different from each other, which conflicts with the phenomenon observed in our proposed framework. Moreover, by leveraging this two-network solution, its performance lags behind ours in both IIW and SAW dataset as shown in Table 3 of the main paper. This performance difference may come from the special design of our AONet that unifies the important features of albedo and shading together, which potentially benefits the albedo and shading prediction from each other.
	
	\begin{figure*}[htp]
		\centering
		\includegraphics[scale=0.47]{analysis_albedo.pdf}\\
		\includegraphics[scale=0.47]{analysis_shading.pdf}
		\vspace{-2mm}
		\caption{Influence of different feature channels to albedo and shading prediction. The error is computed as the difference between the original network output and the one generated by disabling one feature channel of a single layer.}
		\label{figure:analysis}
	\end{figure*}
	\clearpage

	\section{More examples in our dataset, CGIntrinsics \cite{li2018cgintrinsics} and InteriorNet \cite{li2018interiornet} dataset}\label{sec:dataset_comparison}
	
	We further demonstrate more examples from our proposed dataset from Figure \ref{figure:our-data1} to \ref{figure:our-data2}, compared with CGIntrinsics in Figure \ref{figure:CGI-data} and InteriorNet in Figure \ref{figure:Interior-data}. As can be seen, our synthetic data features more complex and realistic lighting environment, richer textures and indoor-scene layouts.
	
	

	\input{figure-our-data1.tex}
	\input{figure-our-data4.tex}
	\input{figure-our-data2.tex}
	\input{figure-CGI-data.tex}
	\input{figure-Interior-data.tex}
	\clearpage

	\section{More qualitative comparisons on the IIW/SAW dataset} \label{sec:results_iiw_saw}
	
	From Figure \ref{figure:iiw_results1} to \ref{figure:iiw_results9}, we demonstrate around 17 groups of visual comparisons with \cite{fan2018revisiting,li2018cgintrinsics} on the IIW/SAW dataset. Our results with ``*'' are generated from the network finetuned on the IIW/SAW dataset.
	
	\input{figure-iiw-results1.tex}
	\input{figure-iiw-results2.tex}
	\input{figure-iiw-results3.tex}
	\input{figure-iiw-results7.tex}
	\input{figure-iiw-results9.tex}
	
	\clearpage

	\section{More visual comparisons on the unseen data source}\label{sec:results_outdoor_scene}
	
	We demonstrate the intrinsic decompositions on around 30 self-collected real-world images which are unseen in the training data from Figure \ref{figure:real_results1} to \ref{figure:real_results12}. Our results are predicted from the network without finetuning on IIW/SAW real data.
	
	\input{figure-real-results1.tex}
	\input{figure-real-results2.tex}
	\input{figure-real-results3.tex}
	\input{figure-real-results16.tex}
	\input{figure-real-results6.tex}
	\input{figure-real-results11.tex}
	\input{figure-real-results4.tex}
	\input{figure-real-results18.tex}
	\input{figure-real-results19.tex}
	\input{figure-real-results12.tex}
	
	\clearpage
	{
		\bibliographystyle{ieee}
		\bibliography{egbib}
	}